\documentclass{article}

\usepackage{microtype}
\usepackage{graphicx}
\usepackage{subcaption}
\usepackage{booktabs}
\usepackage{enumitem}
\usepackage{tikz}
\usetikzlibrary{positioning,arrows.meta,fit,shapes,shapes.geometric,shapes.callouts,calc,backgrounds,decorations.pathreplacing,decorations.pathmorphing,shadows.blur}
\usepackage{hyperref}

\usepackage[accepted]{icml2026}
\usepackage{amsmath}
\usepackage{amssymb}
\usepackage{mathtools}
\usepackage{amsthm}
\usepackage[capitalize,noabbrev]{cleveref}
\usepackage{xcolor}
\usepackage{colortbl}
\usepackage{multirow}
\usepackage{pifont}
\usepackage{fontawesome5}

\theoremstyle{plain}

\theoremstyle{remark}
\newtheorem*{remark}{Remark}

\graphicspath{{figures/}}

\definecolor{teachercol}{HTML}{0072B2}
\definecolor{studentcol}{HTML}{D55E00}
\definecolor{rewardcol}{HTML}{009E73}
\definecolor{stageonebg}{HTML}{EAF1FB}
\definecolor{stagetwobg}{HTML}{FDF2E6}
\icmltitlerunning{WinDOM: Self-Family Distillation}

\begin{document}

\setlength{\textfloatsep}{8pt plus 1pt minus 1pt}
\setlength{\intextsep}{8pt plus 1pt minus 1pt}
\setlength{\abovecaptionskip}{4pt}
\setlength{\belowcaptionskip}{2pt}

\twocolumn[
\icmltitle{WinDOM: Self-Family Distillation for Small-Model GUI Grounding}

\icmlsetsymbol{equal}{*}
\begin{icmlauthorlist}
\icmlauthor{Chengheng Li Chen}{epfl,logitech}
\icmlauthor{Zhiqian Zhou}{epfl}
\icmlauthor{Hao Chen}{upc}
\icmlauthor{Nicolas Chauvin}{logitech}
\end{icmlauthorlist}
\icmlaffiliation{epfl}{\'Ecole Polytechnique F\'ed\'erale de Lausanne (EPFL), Lausanne, Switzerland}
\icmlaffiliation{logitech}{Logitech}
\icmlaffiliation{upc}{Universitat Polit\`ecnica de Catalunya (UPC), Barcelona, Spain}
\icmlcorrespondingauthor{Chengheng Li Chen}{chengheng.lichen@epfl.ch}
\icmlkeywords{GUI grounding, vision-language models, GRPO, self-distillation, synthetic data}
\vskip 0.3in
]

\printAffiliationsAndNotice{}

\begin{abstract}
Small ($\sim$2B) GUI-grounding agents are attractive for on-device deployment, accessibility tooling, and low-cost iteration, but at this scale they face two open recipe questions: how to obtain bounding-box training data without expensive human annotation, and how to combine supervised fine-tuning with reinforcement learning. We address both, with the explicit goal of pushing small-model performance rather than scaling up. \textbf{WinDOM} is a $54{,}425$-record grounding corpus harvested by driving an open-source Windows~11 web reimplementation under headless Playwright, with bounding boxes read directly off the DOM and no OCR or human annotation. \emph{Self-Family Distillation} (SFD) is a single rejection-sampling cold-start parameterised only by the teacher choice: either an EMA of the student (no external model) or a frozen larger same-family teacher. We then treat the saturation depth of the SFD cold-start as an explicit GRPO hyperparameter. On a Qwen3.5-2B student, the under-saturated cold-start is a better GRPO initialiser than the converged one: SFD-4B with Early-init RL gains $+5.4$ OOD-mean ($+3.5$ ScreenSpot-Pro, $+7.0$ OSWorld-G, $+5.8$ ScreenSpot-V2) over the base. The same-size EMA mode lands within roughly one OOD-mean point of the cross-size $4$B variant ($65.2$ vs $66.3$) without an external teacher.
\end{abstract}

\section{Introduction}
\label{sec:intro}

A GUI agent that cannot reliably click the right pixel cannot do anything else: at small ($\sim$2B) scale, grounding accuracy is both the capability bottleneck and the surface on which every downstream tool-call or action-guard ultimately fires. Small-model grounding runs into two costs. \textbf{Data:} professional corpora such as ScreenSpot-Pro~\citep{Cheng2025ScreenSpotPro} need expensive human bounding-box annotation, and OCR-mined alternatives~\citep{Cheng2024SeeClick} inject label noise. \textbf{Post-training:} there is no consensus on how to combine SFT with RL, with recipes split between skipping SFT~\citep{Shao2024DeepSeekMath,HelloKKMe2025GTA1} and launching GRPO from a fully converged SFT base~\citep{InfiGUI2025R1}; parallel evidence~\citep{Liu2025R1Zero,RLHealsOOD2025} points to the \emph{saturation depth} of the SFT initialisation, not its presence, as the real lever. Reliable grounding is also a safety prerequisite for agents in the wild: a model that cannot locate the intended control cannot be constrained to it, so grounding errors surface downstream as clicks on unintended, possibly harmful, UI elements rather than as clean refusals.

\begin{figure*}[t]
\centering
\vspace*{0.65cm}
\resizebox{\textwidth}{!}{%
\begin{tikzpicture}[
  font=\footnotesize\sffamily,
  >={Stealth[length=2.4mm]},
  fwd/.style={->, line width=0.7pt, black!80},
  grad/.style={->, line width=1.5pt, dashed, purple!75!black},
  bridge/.style={->, line width=2.4pt, gray!75!black, line cap=round,
                 >={Stealth[length=3.5mm,width=3mm]}},
  modelcardT/.style={rounded corners=4pt, draw=black!35,
                     line width=0.45pt, fill=white,
                     minimum width=1.9cm, minimum height=1.45cm,
                     inner sep=0pt,
                     blur shadow={shadow blur steps=4, shadow xshift=0.5pt,
                                  shadow yshift=-1.2pt, shadow opacity=22}},
  modelcardS/.style={rounded corners=4pt, draw=black!35,
                     line width=0.45pt, fill=white,
                     minimum width=1.9cm, minimum height=1.45cm,
                     inner sep=0pt,
                     blur shadow={shadow blur steps=4, shadow xshift=0.5pt,
                                  shadow yshift=-1.2pt, shadow opacity=22}},
  cardmarker/.style={line width=8pt, line cap=butt, opacity=0.78},
  klnode/.style={rounded corners=4pt, draw=black!55, fill=orange!12,
                 line width=0.6pt, font=\scriptsize\sffamily,
                 align=center, inner sep=3pt,
                 minimum width=2.05cm, minimum height=1.05cm,
                 blur shadow={shadow blur steps=4, shadow xshift=0.4pt,
                              shadow yshift=-0.5pt, shadow opacity=14}},
  decisiondiamond/.style={diamond, draw=black!70, fill=yellow!22,
                 aspect=1.5, inner sep=1pt,
                 font=\scriptsize\sffamily, align=center,
                 minimum width=2.0cm, minimum height=1.4cm,
                 blur shadow={shadow blur steps=4, shadow xshift=0.4pt,
                              shadow yshift=-0.6pt, shadow opacity=15}},
  rolloutcard/.style={rounded corners=1.5pt, draw=black!50, fill=white,
                      blur shadow={shadow blur steps=4, shadow xshift=0.3pt,
                                   shadow yshift=-0.4pt, shadow opacity=14},
                      minimum width=2.05cm, minimum height=5.5mm,
                      font=\footnotesize\sffamily, inner sep=2pt},
  rewardchipgreen/.style={rounded corners=2pt, draw=rewardcol!55!black,
                      fill=rewardcol!22, line width=0.4pt,
                      minimum width=0.55cm, minimum height=0.42cm,
                      font=\scriptsize\sffamily\bfseries,
                      text=rewardcol!40!black, inner sep=1pt},
  rewardchipred/.style={rounded corners=2pt, draw=red!55!black,
                      fill=red!18, line width=0.4pt,
                      minimum width=0.55cm, minimum height=0.42cm,
                      font=\scriptsize\sffamily\bfseries,
                      text=red!50!black, inner sep=1pt},
  rewardhex/.style={regular polygon, regular polygon sides=6,
                    draw=rewardcol!55!black, line width=0.7pt,
                    fill=rewardcol!18, font=\scriptsize\sffamily,
                    text=rewardcol!35!black, align=center,
                    inner sep=1pt, minimum width=1.5cm},
  advnode/.style={rounded corners=3pt, draw=black!70, fill=gray!10,
                  minimum width=2.5cm, minimum height=1.0cm,
                  font=\footnotesize\sffamily, align=center, inner sep=2pt},
  bestnode/.style={rounded corners=4pt, draw=rewardcol!55!black,
                   fill=rewardcol!12, line width=1.1pt,
                   minimum width=2.6cm, minimum height=1.1cm,
                   font=\footnotesize\sffamily, align=center, inner sep=2pt},
  stagetitle/.style={font=\bfseries\small\sffamily},
  capLbl/.style={font=\scriptsize\sffamily, fill=white,
                 inner xsep=2.5pt, inner ysep=1pt},
  pillLbl/.style={rounded corners=2pt, fill=white, draw=black!25,
                 line width=0.25pt, inner xsep=3pt, inner ysep=1pt,
                 font=\scriptsize\sffamily\bfseries},
  iconsqT/.style={rounded corners=2pt, fill=teachercol!75!black,
                 draw=teachercol!50!black,
                 line width=0.3pt, minimum size=0.42cm, inner sep=0pt,
                 text=white, font=\bfseries\footnotesize},
  iconsqS/.style={rounded corners=2pt, fill=studentcol!75!black,
                 draw=studentcol!50!black,
                 line width=0.3pt, minimum size=0.42cm, inner sep=0pt,
                 text=white, font=\bfseries\footnotesize}
]

\node[fill=stageonebg!55, rounded corners=8pt, draw=black!18, line width=0.4pt,
      minimum width=9.5cm, minimum height=6.10cm, anchor=south west,
      blur shadow={shadow blur steps=4, shadow xshift=0.6pt,
                   shadow yshift=-1.6pt, shadow opacity=22}]
  (stageonepanel) at (-0.25, -3.05) {};

\node[iconsqT, anchor=west]
  (s1icon) at (-0.25, 3.45) {\faSnowflake};
\node[stagetitle, text=teachercol!85!black, anchor=west]
  (s1title) at ([xshift=4pt]s1icon.east)
  {Stage~1: Self-Family Distillation \textit{(cold-start)}};
\draw[teachercol!70!black, line width=0.9pt]
  (-0.25, 3.18) -- (s1title.east |- 0,3.18);

\node[inner sep=0pt, opacity=0.85] (cu2) at (0.65, 1.50)
  {\includegraphics[width=1.30cm]{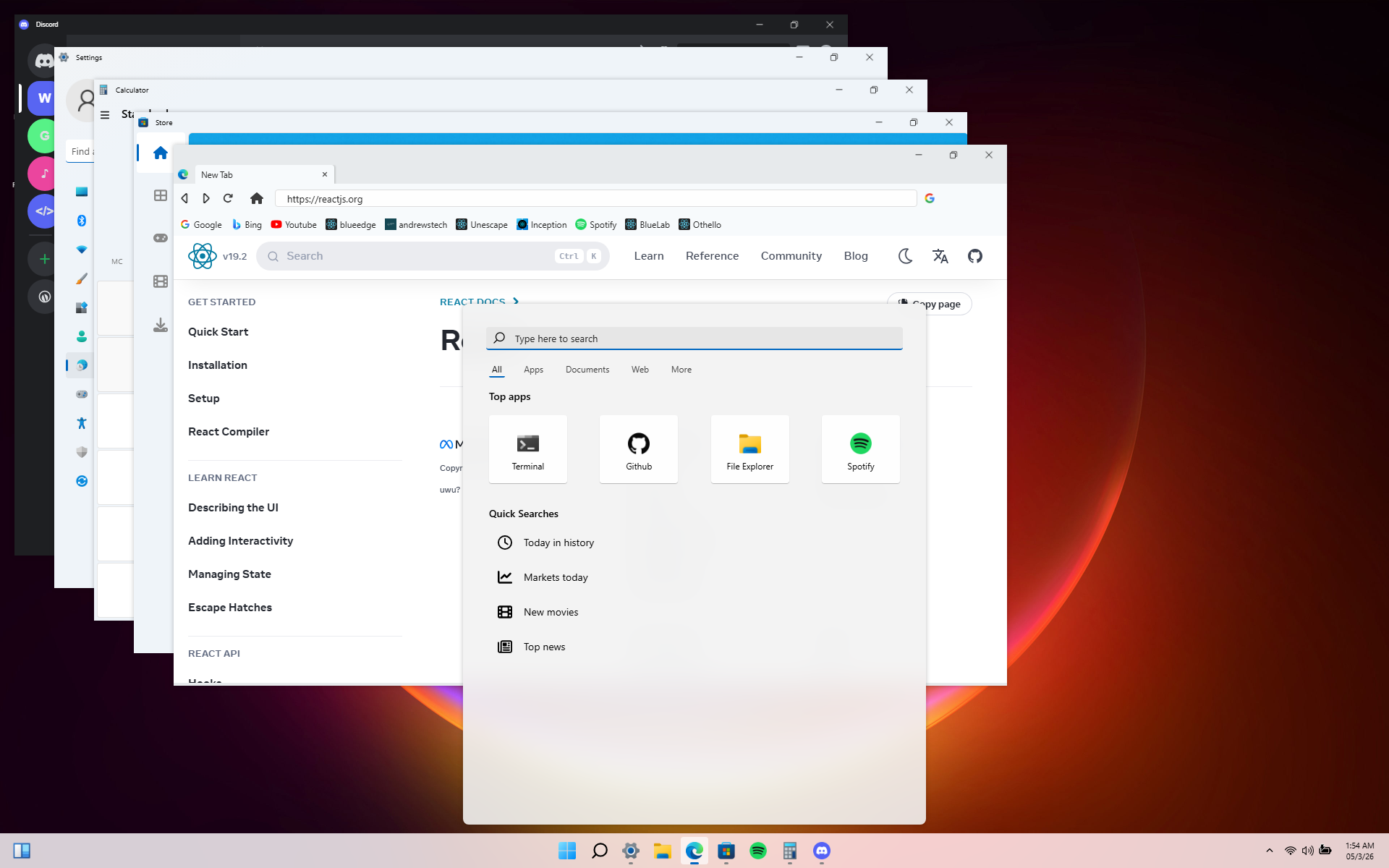}};
\node[inner sep=0pt] (cu1) at (0.85, 1.30)
  {\includegraphics[width=1.30cm]{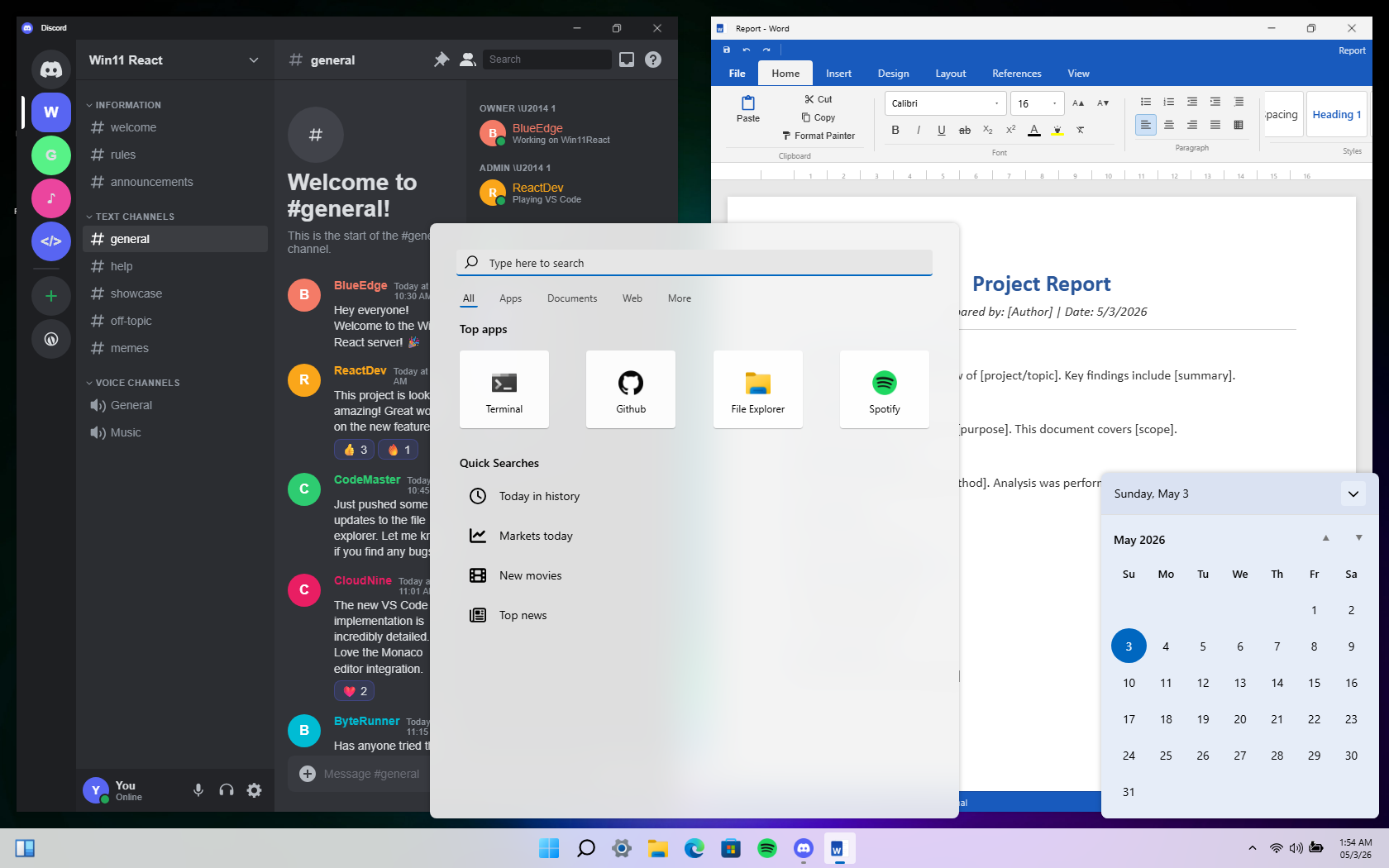}};
\draw[white, line width=1.0pt]
  ([xshift=-0.5pt,yshift=0.5pt]cu1.north west) rectangle
  ([xshift=0.5pt,yshift=-0.5pt]cu1.south east);
\draw[black!55, line width=0.4pt]
  ([xshift=-1.4pt,yshift=1.4pt]cu1.north west) rectangle
  ([xshift=1.4pt,yshift=-1.4pt]cu1.south east);
\draw[orange!85!red, line width=1.4pt, dashed]
  ([xshift=0.20cm,yshift=-0.30cm]cu1.north west) rectangle
  ([xshift=0.60cm,yshift=-0.55cm]cu1.north west);
\node[pillLbl, text=red!55!black, anchor=south west]
  at ([xshift=2pt,yshift=2pt]cu1.south west) {with hint};

\node[inner sep=0pt, opacity=0.85] (cd2) at (0.65, -1.75)
  {\includegraphics[width=1.30cm]{assets/screen_b.png}};
\node[inner sep=0pt] (cd1) at (0.85, -1.95)
  {\includegraphics[width=1.30cm]{assets/screen_a.png}};
\draw[white, line width=1.0pt]
  ([xshift=-0.5pt,yshift=0.5pt]cd1.north west) rectangle
  ([xshift=0.5pt,yshift=-0.5pt]cd1.south east);
\draw[black!55, line width=0.4pt]
  ([xshift=-1.4pt,yshift=1.4pt]cd1.north west) rectangle
  ([xshift=1.4pt,yshift=-1.4pt]cd1.south east);
\node[pillLbl, text=black!65, anchor=south west]
  at ([xshift=2pt,yshift=2pt]cd1.south west) {no hint};

\node[modelcardT] (teacher) at (3.10, 1.30) {};
\draw[cardmarker, teachercol!55]
  ([xshift=-0.65cm,yshift=-0.18cm]teacher.north) --
  ([xshift= 0.65cm,yshift=-0.18cm]teacher.north);
\node[font=\scriptsize\sffamily\bfseries, text=black!90,
      anchor=center] at ([yshift=-0.18cm]teacher.north) {Teacher};
\node[font=\normalsize\sffamily, text=teachercol!75!black,
      anchor=center] at ([yshift=-0.05cm]teacher.center) {$\pi_{\mathrm{T}}$};
\node[font=\scriptsize\sffamily\itshape, text=teachercol!50!black,
      anchor=south] at ([yshift=3pt]teacher.south)
  {\faSnowflake\,frozen};

\node[modelcardS] (student) at (3.10, -1.95) {};
\draw[cardmarker, studentcol!55]
  ([xshift=-0.65cm,yshift=-0.18cm]student.north) --
  ([xshift= 0.65cm,yshift=-0.18cm]student.north);
\node[font=\scriptsize\sffamily\bfseries, text=black!90,
      anchor=center] at ([yshift=-0.18cm]student.north) {Student};
\node[font=\normalsize\sffamily, text=studentcol!75!black,
      anchor=center] at ([yshift=-0.05cm]student.center) {$\pi_{\mathrm{S}}$};
\node[font=\scriptsize\sffamily\itshape, text=studentcol!55!black,
      anchor=south] at ([yshift=3pt]student.south)
  {\faPenNib\,trainable};

\draw[fwd] (cu1.east) -- (teacher.west);
\draw[fwd] (cd1.east) -- (student.west);

\node[decisiondiamond] (rsfilt) at (7.20, 1.30)
  {$\mathbf{y}_{\text{teach}}$\\in bbox?};
\draw[fwd] (teacher.east) -- (rsfilt.west);
\node[font=\tiny\sffamily\itshape, text=teachercol!70!black,
      anchor=north] at ([yshift=-1pt]$ (teacher.east)!0.5!(rsfilt.west) $)
  {$\mathbf{y}_{\text{teach}}$};

\node[klnode] (kl) at (7.20, -0.45)
  {\textbf{Token-KL} $\mathcal{L}_{\mathrm{KD}}$\\[1pt]
   {\fontsize{6.4pt}{7.6pt}\selectfont
    $\mathrm{KL}(\mathbf{y}_{\text{teach}}\,\|\,\mathbf{y}_{\text{student}})$}};

\draw[fwd] (rsfilt.south) -- (kl.north)
  node[pos=0.50, right=2pt, font=\tiny\sffamily\bfseries,
       text=teal!50!black, inner sep=0pt] {yes};

\draw[fwd, dashed, red!70] (rsfilt.north) -- ++(0,0.30)
  node[pos=0.55, right=2pt, font=\tiny\sffamily\bfseries,
       text=red!60!black, inner sep=0pt] {no};
\node (dropglyph) [circle, draw=red!70!black, line width=0.6pt,
              fill=red!8, inner sep=0pt, minimum size=4mm,
              font=\scriptsize\sffamily\bfseries, text=red!70!black]
  at ([yshift=0.55cm]rsfilt.north) {$\oslash$};

\draw[fwd] (student.east) -| (kl.south);
\node[font=\tiny\sffamily\itshape, text=studentcol!70!black,
      anchor=south] at (5.50, -1.95)
  {$\mathbf{y}_{\text{student}}$};

\draw[grad] (kl.west) -| (student.north);
\node[capLbl, text=purple!75!black, font=\scriptsize\sffamily\bfseries]
  at ($ (kl.west)!0.5!(kl.west -| student.north) $)
    {$\nabla\mathcal{L}_{\mathrm{KD}}$};

\node[fill=stagetwobg!55, rounded corners=8pt, draw=black!18, line width=0.4pt,
      minimum width=10.30cm, minimum height=6.10cm, anchor=south west,
      blur shadow={shadow blur steps=4, shadow xshift=0.6pt,
                   shadow yshift=-1.6pt, shadow opacity=22}]
  (stagetwopanel) at (10.30, -3.05) {};

\node[iconsqS, anchor=west]
  (s2icon) at (10.30, 3.45) {\faBolt};
\node[stagetitle, text=studentcol!75!black, anchor=west]
  (s2title) at ([xshift=4pt]s2icon.east)
  {Stage~2: GRPO with verifiable reward};
\draw[studentcol!70!black, line width=0.9pt]
  (10.30, 3.18) -- (s2title.east |- 0,3.18);

\node[modelcardS] (student2) at (11.55, 0.45) {};
\draw[cardmarker, studentcol!55]
  ([xshift=-0.65cm,yshift=-0.18cm]student2.north) --
  ([xshift= 0.65cm,yshift=-0.18cm]student2.north);
\node[font=\scriptsize\sffamily\bfseries, text=black!90,
      anchor=center] at ([yshift=-0.18cm]student2.north) {Student};
\node[font=\normalsize\sffamily, text=studentcol!75!black,
      anchor=center] at ([yshift=-0.05cm]student2.center) {$\pi_{\mathrm{S}}$};
\node[font=\scriptsize\sffamily\itshape, text=studentcol!55!black,
      anchor=south] at ([yshift=3pt]student2.south)
  {\faPenNib\,trainable};


\node[rolloutcard] (r1) at (14.10,  1.35) {$\mathbf{o}^{(1)}$};
\node[rolloutcard] (r2) at (14.10,  0.45) {$\mathbf{o}^{(2)}$};
\begin{scope}[every node/.style={font=\footnotesize\sffamily, text=black!55,
              anchor=center, inner sep=0pt}]
  \node at (14.10, -0.15) {$\cdot$};
  \node at (14.10, -0.35) {$\cdot$};
  \node at (14.10, -0.55) {$\cdot$};
  \node at (14.10, -0.75) {$\cdot$};
\end{scope}
\node[rolloutcard] (r3) at (14.10, -1.15) {$\mathbf{o}^{(G)}$};

\node[font=\scriptsize\sffamily\itshape, text=black!65, anchor=south]
  at (14.10, 1.72) {Sample $G$ rollouts};

\node[rewardchipgreen, minimum width=1.4cm, minimum height=0.55cm,
      font=\scriptsize\sffamily\bfseries, text=rewardcol!35!black]
  (rfn) at (16.50, 0.45) {reward $r$};

\node[rewardhex] (adv) at (19.40, 0.45)
  {Group-rel.\\advantage};

\coordinate (fanjunc) at (12.75, 0.45);
\draw[line width=0.7pt, black!80] (student2.east) -- (fanjunc);
\draw[fwd] (fanjunc) |- (r1.west);
\draw[fwd] (fanjunc) -- (r2.west);
\draw[fwd] (fanjunc) |- (r3.west);

\draw[fwd] (r1.east) -| (rfn.north);
\draw[fwd] (r2.east) -- (rfn.west);
\draw[fwd] (r3.east) -| (rfn.south);

\draw[fwd] (rfn.east) -- (adv.west);

\draw[grad] (adv.south) -- ++(0,-1.50) -| (student2.south);
\coordinate (gradL) at ($ (adv.south) + (0,-1.50) $);
\coordinate (gradR) at (gradL -| student2.south);
\node[pillLbl, text=purple!70!black]
  at ($ (gradL)!0.5!(gradR) $)
  {$\nabla\mathcal{J}_{\mathrm{GRPO}}$};

\end{tikzpicture}%
}
\caption{Pipeline of our best configuration. \textbf{Stage 1 (blue):} a frozen same-family teacher $\pi_{\mathrm{T}}$ receives the screenshot plus a jittered bounding-box hint; the student $\pi_{\mathrm{S}}$ receives only the screenshot. Teacher rollouts that miss the GT box are rejected, and the surviving ones supervise $\pi_{\mathrm{S}}$ via the forward-KL distillation loss $\mathcal{L}_{\mathrm{KD}}$ (\cref{eq:sfd}). \textbf{Stage 2 (orange):} initialised from the step-100 SFD checkpoint, $\pi_{\mathrm{S}}$ samples $G{=}8$ rollouts per prompt; group-relative advantages on a Gaussian click-in-bounding-box reward drive the GRPO update.}
\label{fig:overview}
\end{figure*}

We address both in one pipeline. For data, we render an open-source Windows~11 web clone (Win11React) headlessly and read layout-rectangle bounding boxes straight from the DOM, eliminating OCR and human labels; the corpus, \textbf{WinDOM}, has $500$ trajectories, $2{,}013$ screenshots, and $54{,}425$ grounded records. For post-training, we unify a frozen larger same-family teacher (cross-size) and an EMA self-teacher (same-size)~\citep{Shenfeld2026SelfDistillContinual} under one click-in-bounding-box rejection-sampling cold-start we call \emph{Self-Family Distillation} (SFD), and treat its \emph{saturation depth} as the central GRPO hyperparameter (\cref{fig:overview}: SFD-4B, then GRPO from an under-saturated step-$100$ checkpoint). Relevant to deploying agents in the wild, every supervisory signal here, the SFD filter and the GRPO reward, is a single deterministic point-in-box check rather than a learned reward model or LLM judge, so the whole training channel is cheap to audit and reproduce.

\paragraph{Contributions.}
\begin{enumerate}[leftmargin=*,topsep=0pt,itemsep=0pt]
\item \textbf{WinDOM.} A 54k-record GUI grounding corpus with bounding boxes read directly from the DOM layout rectangle (no OCR, no human annotation), harvested from an open-source Windows~11 clone and released with a deterministic regeneration script.
\item \textbf{Self-Family Distillation.} One rejection-sampled cold-start unifying an EMA self-teacher~\citep{Shenfeld2026SelfDistillContinual} and a frozen larger same-family teacher, set by the teacher choice alone; the teacher-free EMA variant comes within one OOD-mean point of the 4B teacher.
\item \textbf{Cold-start depth as a GRPO hyperparameter.} Early-init RL on the SFD-4B cold-start trades a few in-distribution points for $+3.5$ ScreenSpot-Pro, $+7.0$ OSWorld-G, and $+5.8$ ScreenSpot-V2 over the Qwen3.5-2B base; Late-init RL holds in-distribution accuracy but barely moves OOD in every regime.
\end{enumerate}

\section{Related Work}
\label{sec:related}

We situate WinDOM and Self-Family Distillation against three lines of prior work: GUI-grounding corpora and small-model grounding agents, filtered teacher-student distillation in language modelling, and reinforcement learning from cold-started supervised checkpoints.

\paragraph{GUI grounding datasets and models.}
The standard evaluation suites for click-level grounding are ScreenSpot-Pro~\citep{Cheng2025ScreenSpotPro}, OSWorld-G~\citep{Xie2025Jedi}, and ScreenSpot-V2~\citep{Wu2024OSAtlas}; on the training-corpus side, OS-Atlas~\citep{Wu2024OSAtlas} and Jedi~\citep{Xie2025Jedi} reach desktop scale by mining or decomposing real screenshots. WinDOM is desktop-OS-specific, reads layout-rectangle bounding boxes directly from the DOM of a deterministic Win11React renderer, and is seedable end-to-end so the corpus can be regenerated from a single seed. Our backbone is Qwen3.5-2B~\citep{Qwen2025Qwen3VL}, whose tool-calling interface emits coordinates inside a structured JSON action and removes the need for an auxiliary grounding head.

\paragraph{Reinforcement learning and cold-start depth.}
Recent GUI-RL recipes operate at $\geq 3$B and either skip the cold-start~\citep{HelloKKMe2025GTA1} or initialise from a single fully converged supervised checkpoint~\citep{UIR12025,Zhou2025GUIG1,UITARS22025}, treating that choice as fixed. Parallel evidence in the reasoning literature~\citep{Chu2025SFTMemRL,RLHealsOOD2025} shows that out-of-distribution accuracy peaks early in supervised training and that subsequent reinforcement learning can recover part of what supervised training has lost. We extend that observation to GUI grounding at 2B and turn \emph{cold-start depth} into an explicitly ablated hyperparameter, mapping its effect on the in-distribution / out-of-distribution Pareto frontier.

\paragraph{Self-distillation and filtered training signals.}
SFD draws on two established lines of prior work. The same-size mode inherits from EMA self-teaching, introduced for representation learning by Mean Teacher~\citep{Tarvainen2017MeanTeacher} and BYOL~\citep{Grill2020BYOL} and for same-architecture distillation by Born-Again Networks~\citep{Furlanello2018BAN}; SDPO~\citep{SDPO2026}, Metis-SPECS~\citep{Metis2025}, and the continual-learning self-distillation analysis of \citet{Shenfeld2026SelfDistillContinual} carry the same idea into RL and lifelong-learning pipelines. The cross-size mode is a filtered, off-policy variant of the same-family-teacher distillation popularised for reasoning by ReST$^{\mathrm{EM}}$~\citep{Singh2023ReSTEM}, RAFT, and STaR~\citep{Zelikman2022STaR}, with the learned or LLM-judged verifier of those methods replaced by a deterministic geometric click-in-bounding-box check. Our contribution is methodological rather than algorithmic: both modes share a single rejection-sampling training script parameterised only by the teacher choice, and the resulting checkpoint is taken under-saturated by design to serve as a cold-start for downstream GRPO rather than as a terminal stage.
\section{Problem Definition}
\label{sec:background}

We formalise grounding as click prediction on (screenshot, instruction, bounding-box) triples, and fix the policy class and accuracy metric used throughout the paper.

\subsection{The grounding task}
\label{sec:task}
A grounding instance is a triple
\begin{align}
&(\mathbf{x}_v,\,\mathbf{x}_t,\,\mathcal{B}_{\text{GT}}), \nonumber \\
&\mathbf{x}_v \in \mathbb{R}^{H\times W\times 3}, \quad \mathcal{B}_{\text{GT}}\subset[0,W]\times[0,H],
\label{eq:instance}
\end{align}
in which $\mathbf{x}_v$ is a screenshot, $\mathbf{x}_t$ is a free-form natural-language instruction, and $\mathcal{B}_{\text{GT}}$ is the axis-aligned bounding box of the screen element that $\mathbf{x}_t$ describes. The task is to map the prompt $\mathbf{x}=(\mathbf{x}_v,\mathbf{x}_t)$ to a click
\begin{equation}
\mathbf{p}\in[0,W]\times[0,H], \quad \mathbf{p}\in\mathcal{B}_{\text{GT}}.
\label{eq:click}
\end{equation}
We do not assume that the instruction uniquely identifies a single visual element; when an instance admits multiple plausible targets, the model is supervised and judged only against the one box that ships with the instance.

\subsection{Model class}
\label{sec:model-class}
A grounding model is a vision-language policy
\begin{equation}
\pi_\theta(\mathbf{o}\mid\mathbf{x}), \qquad \mathbf{o}=(o_1,\dots,o_T)\in\mathcal{V}^*,
\label{eq:policy}
\end{equation}
paired with a fixed parser
\begin{equation}
\mathrm{decode}:\mathcal{V}^* \to \big([0,W]\times[0,H]\big)\cup\{\bot\}
\label{eq:decode}
\end{equation}
that extracts a click candidate from $\mathbf{o}$ and returns $\bot$ when the output fails to parse. The parser is part of the environment rather than the model: a non-parsable rollout counts as an incorrect click, not as a separate failure mode. In our concrete instantiation $\pi_\theta$ is a tool-using vision-language transformer that emits integer coordinates on a $[0,1000]^2$ scale, rescaled by $(W,H)$ before scoring; \cref{sec:experiments} reports the exact emission and parsing protocol.

\subsection{Click-in-target accuracy}
\label{sec:metric}
The \emph{click-in-target} indicator on an output $\mathbf{o}$ is
\begin{equation}
\mathrm{CIT}(\mathbf{o},\mathcal{B}_{\text{GT}}) \;\;\coloneqq\;\; \mathbf{1}\!\left[\mathrm{decode}(\mathbf{o}) \in \mathcal{B}_{\text{GT}}\right],
\label{eq:cgt}
\end{equation}
and click-in-target accuracy on a benchmark $\mathcal{D}$ is its empirical mean under greedy decoding,
\begin{align}
\mathrm{Acc}(\mathcal{D}) &= \mathbb{E}_{(\mathbf{x},\mathcal{B}_{\text{GT}})\sim\mathcal{D}}\bigl[\mathrm{CIT}(\hat{\mathbf{o}},\mathcal{B}_{\text{GT}})\bigr], \nonumber \\
\hat{\mathbf{o}} &= \arg\max_{\mathbf{o}}\pi_\theta(\mathbf{o}\mid\mathbf{x}).
\label{eq:acc}
\end{align}
This is the only metric reported in this paper. It is strictly stricter than center-distance and strictly looser than full bounding-box localisation: a click is correct iff it falls inside the rectangle, regardless of its distance to the centre, and the model is never required to recover the rectangle itself.

\section{Data Collection: WinDOM}
\label{sec:dataset}

We construct WinDOM by treating an open-source browser reimplementation of Windows~11 as a self-annotating screenshot generator, recording paired (screenshot, DOM) snapshots from which bounding boxes are read directly off the document layout.

\paragraph{Why a browser?} Operating systems are difficult to introspect: screenshots require a virtual machine, labels come from OCR or human annotation, and visual elements have no clean correspondence with code-level identifiers. A browser-based reimplementation reverses these properties: every visible element is a DOM node whose bounding box is exposed to JavaScript, whose accessibility role is stored in the document, and whose developer-supplied attributes already name the underlying action. The browser is, in effect, a self-annotating screenshot generator.

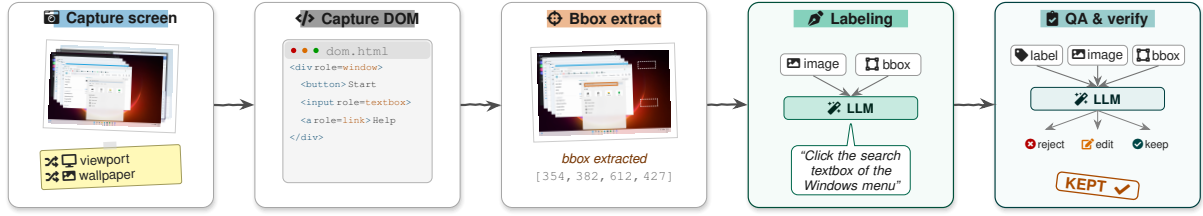
\begin{figure*}[t]
\centering
\resizebox{0.92\textwidth}{!}{%
\begin{tikzpicture}[
  font=\footnotesize\sffamily,
  >={Stealth[length=2.6mm]},
  fwd/.style={-{Stealth[length=2.4mm]}, line width=1.0pt, black!70,
              decorate, decoration={random steps, segment length=2.2pt, amplitude=0.30pt}},
  frame/.style={rounded corners=5pt, draw=black!35, line width=0.45pt,
                fill=white, minimum width=3.30cm, minimum height=3.30cm,
                inner sep=0pt, blur shadow={shadow blur steps=4,
                shadow xshift=0.5pt, shadow yshift=-1.4pt, shadow opacity=22}},
  marker/.style={line width=8pt, line cap=butt, opacity=0.78},
  titletext/.style={font=\scriptsize\sffamily\bfseries, text=black!90,
                    anchor=center, inner sep=0pt},
  shotborder/.style={black!55, line width=0.35pt},
  stickynote/.style={fill=yellow!35, draw=yellow!55!black, line width=0.3pt,
                     rounded corners=1pt, font=\scriptsize\sffamily,
                     inner sep=2.2pt, text=black!85}
]

\def\framew{2.80cm}
\def\frameh{3.75cm}
\def\framegap{0.65cm}
\def\shotw{1.95cm}
\def\barw{2.40cm}
\tikzset{arrowlabel/.style={font=\tiny\sffamily, text=black!65,
                            inner sep=1pt, fill=white,
                            rounded corners=1pt, midway, above=-0.5pt}}

\node[frame] (f1) at (0,0) {};
\draw[marker, teachercol!55]
  ([xshift=-0.92cm,yshift=-0.27cm]f1.north) -- ([xshift=0.92cm,yshift=-0.27cm]f1.north);
\node[titletext, font=\scriptsize\sffamily\bfseries]
  at ([yshift=-0.27cm]f1.north) {\faCameraRetro\;\,Capture screen};

\node[draw=black!20, line width=0.4pt, fill=white, inner sep=2.5pt,
      rotate=-2.2, anchor=center]
  (poly1) at ([yshift=+0.30cm]f1.center)
  {\includegraphics[width=1.95cm]{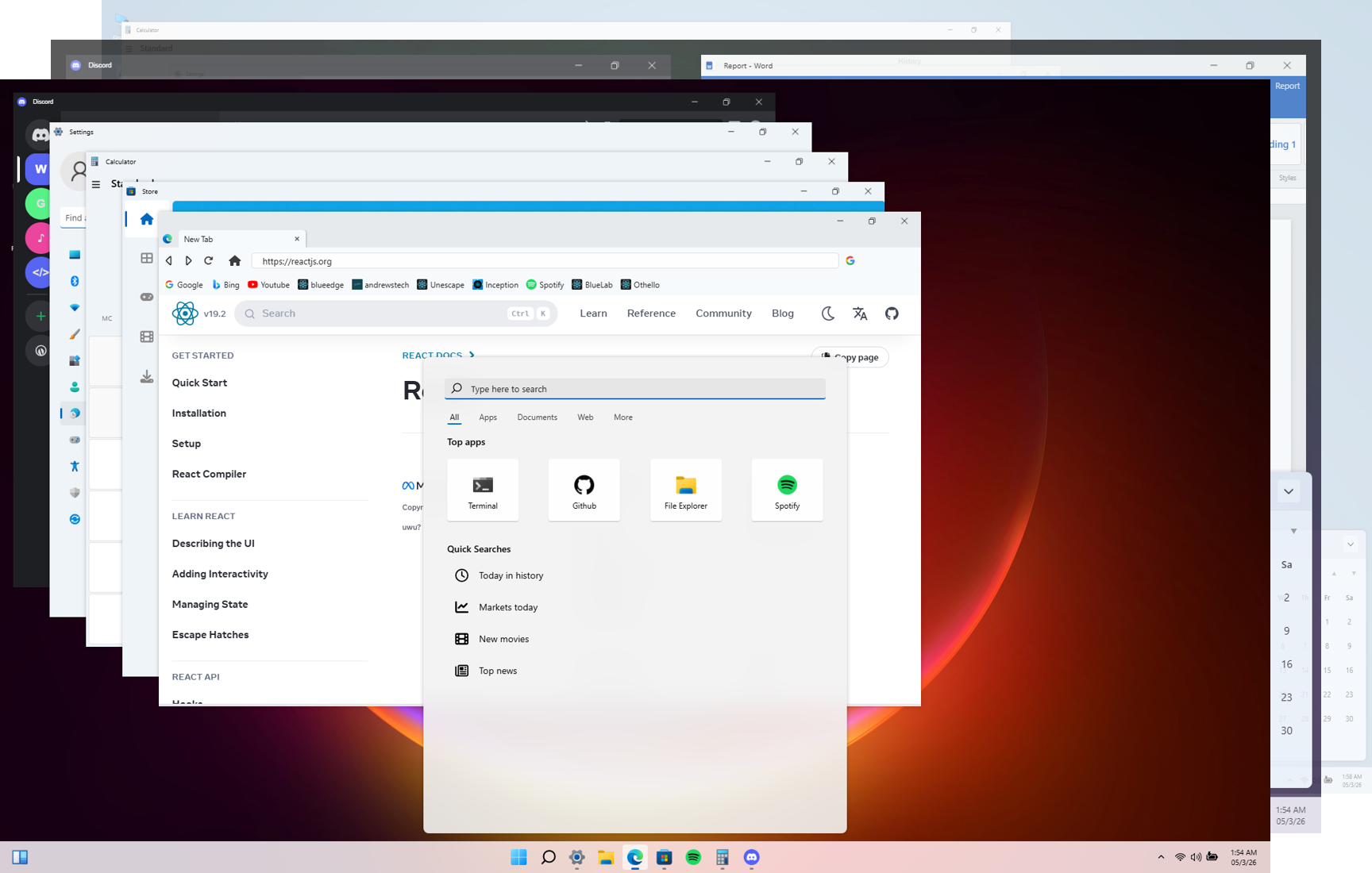}};

\node[stickynote, anchor=center, rotate=2.5, text width=2.10cm, align=left,
      font=\tiny\sffamily]
  (note1) at ([yshift=-1.00cm]f1.center)
  {\faRandom\;\faDesktop\;viewport\\\faRandom\;\faImage\;wallpaper};
\fill[black!18, opacity=0.55, rotate around={2.5:(note1.north)}]
  ([xshift=-3pt,yshift=-1pt]note1.north) rectangle
  ([xshift= 3pt,yshift= 2.5pt]note1.north);

\node[frame, right=\framegap of f1] (f2) {};
\draw[marker, black!50]
  ([xshift=-0.92cm,yshift=-0.27cm]f2.north) -- ([xshift=0.92cm,yshift=-0.27cm]f2.north);
\node[titletext, font=\scriptsize\sffamily\bfseries]
  at ([yshift=-0.27cm]f2.north) {\faCode\;\,Capture DOM};

\node[draw=black!30, fill=black!4, rounded corners=2.5pt,
      line width=0.4pt, minimum width=2.40cm, minimum height=2.30cm,
      anchor=center]
  (codepane) at ([yshift=-0.10cm]f2.center) {};
\fill[black!12, rounded corners=2pt]
  ([xshift=1pt,yshift=-1pt]codepane.north west) rectangle
  ([xshift=-1pt,yshift=-9pt]codepane.north east);
\fill[red!75!black]    ([xshift=0.18cm,yshift=-0.18cm]codepane.north west) circle (1.2pt);
\fill[orange!85!black] ([xshift=0.36cm,yshift=-0.18cm]codepane.north west) circle (1.2pt);
\fill[green!60!black]  ([xshift=0.54cm,yshift=-0.18cm]codepane.north west) circle (1.2pt);
\node[font=\tiny\ttfamily, text=black!55, anchor=center]
  at ([yshift=-0.18cm]codepane.north) {dom.html};
\begin{scope}
  \clip[rounded corners=2pt]
    ([xshift=2pt,yshift=-2pt]codepane.north west) rectangle
    ([xshift=-2pt,yshift=2pt]codepane.south east);
  \begin{scope}[every node/.style={font=\fontsize{4.4pt}{5.0pt}\selectfont\ttfamily,
                text=black!85, anchor=north west, align=left, inner sep=0pt}]
    \node at ([xshift=3pt,yshift=-11pt]codepane.north west)
      {\textcolor{teachercol!60!black}{<div}\,role=\textcolor{studentcol!80!black}{window}\textcolor{teachercol!60!black}{>}};
    \node at ([xshift=8pt,yshift=-19pt]codepane.north west)
      {\textcolor{teachercol!60!black}{<button>}\,Start};
    \node at ([xshift=8pt,yshift=-27pt]codepane.north west)
      {\textcolor{teachercol!60!black}{<input}\,role=\textcolor{studentcol!80!black}{textbox}\textcolor{teachercol!60!black}{>}};
    \node at ([xshift=8pt,yshift=-35pt]codepane.north west)
      {\textcolor{teachercol!60!black}{<a}\,role=\textcolor{studentcol!80!black}{link}\textcolor{teachercol!60!black}{>}\,Help};
    \node at ([xshift=3pt,yshift=-43pt]codepane.north west)
      {\textcolor{teachercol!60!black}{</div>}};
  \end{scope}
\end{scope}

\node[frame, right=\framegap of f2] (f3) {};
\draw[marker, studentcol!55]
  ([xshift=-0.92cm,yshift=-0.27cm]f3.north) -- ([xshift=0.92cm,yshift=-0.27cm]f3.north);
\node[titletext, font=\scriptsize\sffamily\bfseries]
  at ([yshift=-0.27cm]f3.north) {\faCrosshairs\;\,Bbox extract};

\node[draw=black!20, line width=0.4pt, fill=white, inner sep=3pt,
      rotate=1.8, anchor=center]
  (poly3) at ([yshift=+0.20cm]f3.center)
  {\includegraphics[width=2.10cm]{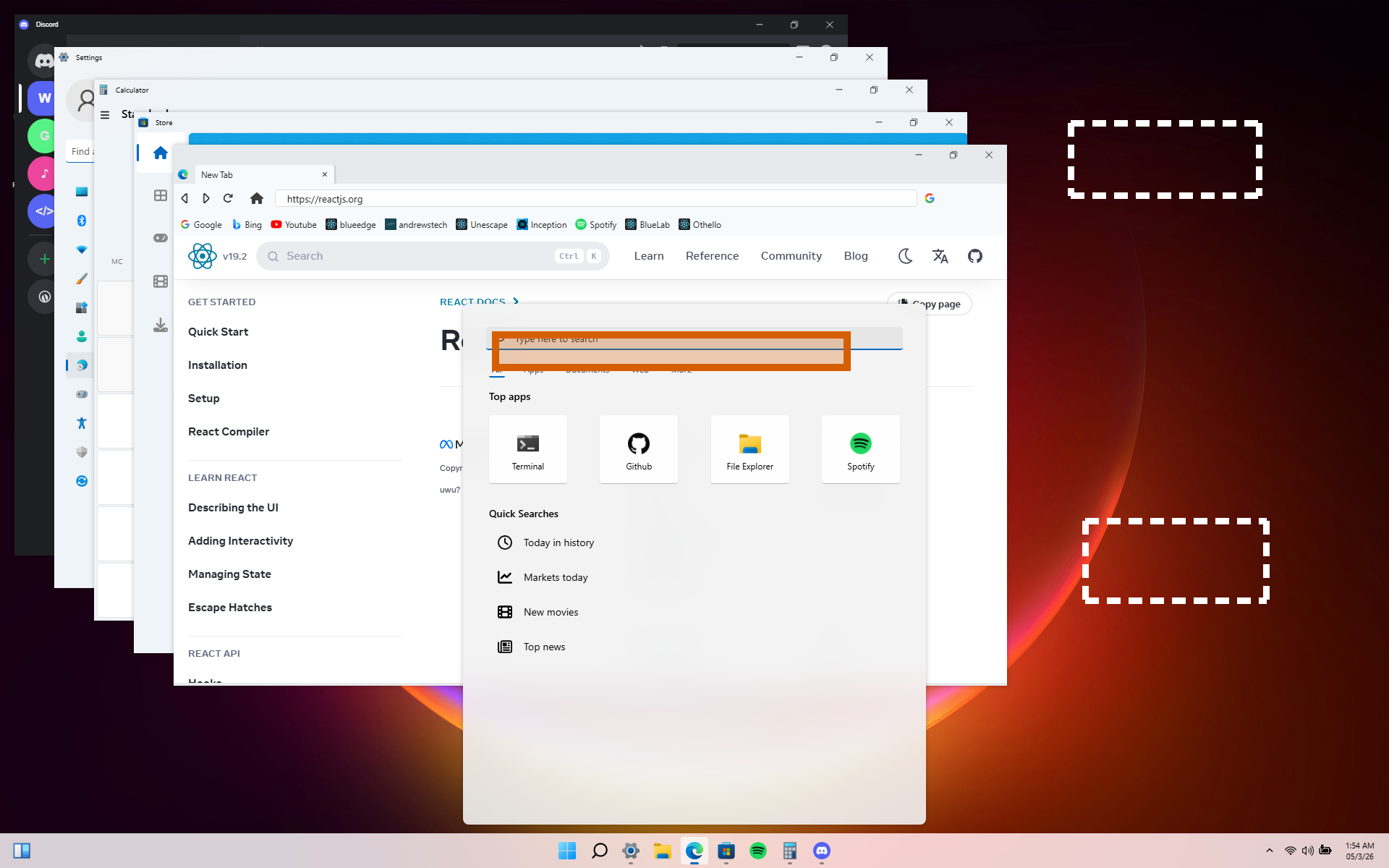}};
\node[font=\tiny\sffamily\itshape, text=studentcol!55!black, anchor=north]
  at ([yshift=-0.65cm]f3.center) {bbox extracted};
\node[font=\tiny\sffamily, text=black!60, anchor=north]
  at ([yshift=-0.95cm]f3.center) {\texttt{[354,\,382,\,612,\,427]}};

\node[frame, right=\framegap of f3, draw=rewardcol!50!black, line width=0.55pt,
      fill=rewardcol!4] (f4) {};
\draw[marker, rewardcol!60]
  ([xshift=-0.92cm,yshift=-0.27cm]f4.north) -- ([xshift=0.92cm,yshift=-0.27cm]f4.north);
\node[titletext, font=\scriptsize\sffamily\bfseries]
  at ([yshift=-0.27cm]f4.north) {\faPenNib\;\,Labeling};

\node[draw=black!40, fill=white, rounded corners=2pt,
      font=\tiny\sffamily, text=black!85, inner xsep=3pt, inner ysep=2pt,
      minimum width=1.05cm, anchor=east]
  (in_img) at ([xshift=-2pt,yshift=+0.65cm]f4.center)
  {\faImage\;image};
\node[draw=black!40, fill=white, rounded corners=2pt,
      font=\tiny\sffamily, text=black!85, inner xsep=3pt, inner ysep=2pt,
      minimum width=1.05cm, anchor=west]
  (in_bbox) at ([xshift=2pt,yshift=+0.65cm]f4.center)
  {\faVectorSquare\;bbox};
\node[draw=rewardcol!65!black, fill=rewardcol!22, rounded corners=2pt,
      font=\tiny\sffamily\bfseries, text=rewardcol!25!black,
      inner sep=2.2pt, anchor=center, minimum width=2.15cm]
  (refine) at ([yshift=-0.05cm]f4.center)
  {\faMagic\;\,LLM};
\draw[->,>=Stealth, line width=0.5pt, black!55] (in_img.south)  -- (refine.north);
\draw[->,>=Stealth, line width=0.5pt, black!55] (in_bbox.south) -- (refine.north);
\node[rectangle callout, rounded corners=3pt,
      draw=rewardcol!65!black, fill=white, line width=0.45pt,
      callout absolute pointer={(refine.south)},
      callout pointer width=3.5pt, callout pointer shorten=0pt,
      inner xsep=3pt, inner ysep=2pt,
      font=\tiny\sffamily\itshape, text=black!90,
      text width=1.70cm, align=center]
  (instr) at ([yshift=-1.05cm]f4.center)
  {``Click the search textbox of the Windows menu''};

\node[frame, right=\framegap of f4, draw=teal!50!black, line width=0.55pt,
      fill=teal!3] (f5) {};
\draw[marker, teal!50]
  ([xshift=-0.92cm,yshift=-0.27cm]f5.north) -- ([xshift=0.92cm,yshift=-0.27cm]f5.north);
\node[titletext, font=\scriptsize\sffamily\bfseries]
  at ([yshift=-0.27cm]f5.north) {\faClipboardCheck\;\,QA \& verify};

\begin{scope}[every node/.style={draw=black!40, fill=white, rounded corners=2pt,
              font=\tiny\sffamily, text=black!85, inner xsep=2pt, inner ysep=1.6pt,
              minimum width=0.72cm, anchor=center, align=center}]
  \node (qain_lab) at ([xshift=-0.98cm,yshift=+0.80cm]f5.center) {\faTag\,label};
  \node (qain_img) at ([xshift= 0.00cm,yshift=+0.80cm]f5.center) {\faImage\,image};
  \node (qain_box) at ([xshift=+0.98cm,yshift=+0.80cm]f5.center) {\faVectorSquare\,bbox};
\end{scope}
\node[draw=teal!50!black, fill=teal!10, rounded corners=2pt,
      font=\tiny\sffamily\bfseries, text=teal!30!black,
      inner sep=2.2pt, anchor=center, minimum width=2.15cm]
  (qajudge) at ([yshift=+0.10cm]f5.center)
  {\faMagic\;\,LLM};
\draw[->,>=Stealth, line width=0.5pt, black!55] (qain_lab.south) -- (qajudge.north);
\draw[->,>=Stealth, line width=0.5pt, black!55] (qain_img.south) -- (qajudge.north);
\draw[->,>=Stealth, line width=0.5pt, black!55] (qain_box.south) -- (qajudge.north);
\draw[->,>=Stealth, line width=0.4pt, black!50] (qajudge.south) --
  ([xshift=-0.85cm,yshift=-0.40cm]f5.center);
\draw[->,>=Stealth, line width=0.4pt, black!50] (qajudge.south) --
  ([xshift= 0.00cm,yshift=-0.40cm]f5.center);
\draw[->,>=Stealth, line width=0.4pt, black!50] (qajudge.south) --
  ([xshift= 0.85cm,yshift=-0.40cm]f5.center);
\begin{scope}[every node/.style={font=\fontsize{5pt}{6pt}\selectfont\sffamily,
              text=black!85, anchor=north, inner sep=1pt}]
  \node at ([xshift=-0.85cm,yshift=-0.50cm]f5.center)
    {\textcolor{red!70!black}{\faTimesCircle}\,reject};
  \node at ([xshift= 0.00cm,yshift=-0.50cm]f5.center)
    {\textcolor{orange!85!black}{\faEdit}\,edit};
  \node at ([xshift= 0.85cm,yshift=-0.50cm]f5.center)
    {\textcolor{teal!55!black}{\faCheckCircle}\,keep};
\end{scope}
\node[draw=studentcol!80!black, line width=0.6pt, rounded corners=2pt,
      font=\scriptsize\sffamily\bfseries, text=studentcol!80!black,
      inner xsep=4pt, inner ysep=1.6pt,
      rotate=-7, anchor=center]
  (stamp) at ([yshift=-1.30cm]f5.center)
  {KEPT \;\faCheck};

\draw[fwd] (f1.east) to[bend left=1] (f2.west);
\draw[fwd] (f2.east) to[bend left=1] (f3.west);
\draw[fwd] (f3.east) to[bend left=1] (f4.west);
\draw[fwd] (f4.east) to[bend left=1] (f5.west);

\end{tikzpicture}%
}
\caption{\textbf{WinDOM data pipeline.} A scene script drives Win11React under headless Playwright with randomized viewport, wallpaper, and layout (\emph{capture screen}). At every step we record the live DOM (\emph{capture DOM}) and extract bounding boxes either from the latest user-action locator or from interactable ARIA roles (\emph{bbox extract}). An LLM emits a short imperative instruction for each crop (\emph{labeling}); a second pass classifies the candidate as \emph{keep}, \emph{edit}, or \emph{reject} (\emph{QA \& verify}). Each kept pass yields one WinDOM record (image, instruction, bbox) with provenance flags.}
\label{fig:dataset}
\end{figure*}

\paragraph{Pipeline.} We instantiate the principle through Win11React, a community-maintained browser reimplementation of Windows~11 driven headlessly under Playwright. The methodology has five stages whose names match the panel titles of \cref{fig:dataset}; \cref{app:dataset} lists the concrete thresholds and parameter values.

\emph{(i)~Capture screen.} A scene is a short script of GUI primitives authored against CSS selectors, rendered under a randomised viewport, wallpaper, and one of six layout templates per scene so the same applications appear under varied visual conditions.

\emph{(ii)~Capture DOM.} At every step we record a paired (screenshot, DOM) snapshot.

\emph{(iii)~Bounding-box extraction.} A single screenshot yields several grounding tasks: one record from the locator that produced the latest user action, plus one record per visible element whose ARIA role belongs to a fixed set of interactable roles. Each box is read directly from the document layout, so the corpus needs neither OCR nor human annotation; the layout rectangle equals the visually-clickable region only modulo CSS shadows, focus outlines, and transformed elements (see \cref{sec:limitations}).

\emph{(iv)~Labelling.} For each extracted bounding box we prompt a small vision-language model (Gemini~2.5~Flash) with the cropped screenshot together with the element's ARIA role, accessible name, and action label, and ask it to emit a short imperative grounding instruction. Conditioning the labeller on both the image and the structured DOM signals reduces the role-name ambiguity that pure-text labelling pipelines suffer from.

\emph{(v)~QA and verification.} A second vision-language pass receives the screenshot, the candidate instruction, and the bounding box, and classifies the candidate as \emph{keep}, \emph{edit}, or \emph{reject}; a subset of \emph{edit} candidates is refined by an automated vision agent and a small fraction is reviewed by a human. Geometric cleaning runs alongside the QA pass to drop degenerate boxes and near-duplicates, and every retained record carries provenance flags marking which stages it has passed.

\paragraph{Output and splits.} The release contains 500 trajectories, 2{,}013 screenshots, and 54{,}425 grounded records, partitioned at the trajectory level into training, validation, and test splits. Training in \cref{sec:experiments} consumes the training partition; further details on the partitioning and the optional cleaned release are deferred to \cref{app:dataset}.

\section{Training Methodology}
\label{sec:method}

The training pipeline has three components: a vision-language backbone (\cref{sec:backbone}), three supervised cold-start regimes (\cref{sec:sft,sec:sfd}), and a GRPO stage initialised from one of the resulting checkpoints (\cref{sec:rl}). We call a cold-start \emph{saturated} when its in-distribution training-accuracy curve has plateaued and \emph{early} when taken at a pre-plateau step; \emph{Late-init RL} and \emph{Early-init RL} denote GRPO runs that share every reinforcement-learning hyperparameter and differ only in that choice.

\subsection{Backbone assumptions}
\label{sec:backbone}
The methodology assumes a vision-language policy $\pi_\theta(\mathbf{o}\mid\mathbf{x})$ in the sense of \cref{sec:background}, with the additional assumption that its output sequence $\mathbf{o}$ contains a structured action sub-sequence $\mathbf{y}\subseteq\mathbf{o}$ whose tokens encode the predicted click coordinate, parsable by a fixed regular expression into a point in the same coordinate frame as $\mathcal{B}_{\text{GT}}$. We write $\mathbf{y}_{\text{teach}}$ and $\mathbf{y}_{\text{student}}$ for the action sub-sequences emitted by the teacher and student in \cref{sec:sfd}, both lying inside their respective $\mathbf{o}$. The backbone is otherwise treated as a black box: no architectural modification, no vocabulary expansion, no auxiliary heads. Specific choices of backbone, parameter count, and tool-call schema are deferred to \cref{sec:experiments}.

\subsection{Cold-start regimes}
\label{sec:sft}
We use three cold-start regimes, each implemented in the same training loop: Direct SFT (this subsection), and two modes of Self-Family Distillation (\cref{sec:sfd}). Direct SFT serves as the baseline foil and is the most common cold-start in the GUI grounding literature~\citep{Cheng2024SeeClick,Hong2024CogAgent}: given a training instance $(\mathbf{x}_v,\mathbf{x}_t,\mathcal{B}_{\text{GT}})$, the target sequence $\mathbf{y}$ is the tokenized action that emits the pixel-space center of $\mathcal{B}_{\text{GT}}$, and the loss is the standard token-level cross-entropy
\begin{equation}
\mathcal{L}_{\text{SFT}}(\theta) \;=\; -\,\mathbb{E}_{(\mathbf{x},\mathbf{y})\sim\mathcal{D}}\sum_{t=1}^{|\mathbf{y}|}\log \pi_\theta(y_t\mid\mathbf{x},\mathbf{y}_{<t}),
\end{equation}
where $\mathcal{D}$ is the WinDOM training distribution. Because the surrounding tokens are a fixed format wrapper, most of the loss mass is concentrated on the two coordinate digits.

Training is checkpointed at uniform intervals so that the cold-start can later be selected at an early or saturated point. The training-set click-in-target accuracy is a unimodal saturating curve in the training step; we treat the plateau as the saturated checkpoint and any pre-plateau point as an early checkpoint.

\subsection{Self-Family Distillation}
\label{sec:sfd}

Beyond Direct SFT, our second cold-start regime is a single algorithm that we call \emph{Self-Family Distillation} (SFD). It is filtered, off-policy distillation onto the student, parameterized by which model serves as the teacher. SFD has two natural modes, \emph{same-size} (the teacher is an EMA of the student; no external model needed) and \emph{cross-size} (the teacher is a frozen, stronger same-family model that shares the student's tokenizer); both use the same loss and training script.

\paragraph{Same-size mode (no external teacher).} The teacher is an exponential moving average (EMA) of the student itself,
\begin{equation}
\theta_{\text{EMA}} \;\leftarrow\; \alpha\,\theta_{\text{EMA}} + (1-\alpha)\,\theta,
\quad \alpha \in (0,1),
\end{equation}
which receives no gradient.

\paragraph{Cross-size mode (larger teacher).} The teacher is a stronger frozen model from the same architectural family as the student. The two checkpoints are required to share the tokenizer and the special-token table so that the teacher's logits over the vocabulary can be compared to the student's without alignment. The specific student/teacher pair is deferred to \cref{sec:experiments}.

\paragraph{Teacher input.} In both modes the teacher receives the screenshot plus a textual \emph{hint} consisting of the ground-truth bounding box jittered by independent integer noise of $\pm 30$ on the $[0,1000]^2$ scale, clamped to range; the hint is added only to the teacher's user prompt and the student is never shown it. The hint is shared by the same-size and cross-size modes so that the rejection-sampling filter (below) acts on the same teacher conditioning regardless of teacher choice; the exact prompt format is reproduced in \cref{app:prompts}.

\paragraph{Filtered teacher rollouts.} In both modes the teacher first generates a candidate completion $\mathbf{y}_{\text{teach}}$ from the hint-conditioned prompt. We parse a coordinate from $\mathbf{y}_{\text{teach}}$ and accept the rollout only if its center lies inside the ground-truth bounding box; otherwise the example is skipped for the current step. The filter prevents distillation from spending student capacity on incorrect teacher rollouts; the empirical kept-rate of each mode is reported in \cref{sec:finding3}.

\paragraph{Loss.} Conditioned on a filtered teacher rollout $\mathbf{y}_{\text{teach}}=(y_1,\dots,y_{|\mathbf{y}_{\text{teach}}|})$, we use the standard token-level distillation loss in cross-entropy form: forward KL from the teacher to the student, evaluated at the positions of $\mathbf{y}_{\text{teach}}$. Letting $\pi^{(t)}(\cdot)\equiv\pi(\cdot\mid\mathbf{x},\mathbf{y}_{\text{teach},<t})$,
\begin{equation}
\mathcal{L}_{\text{SFD}}(\theta)
\;=\; \mathbb{E}_{\mathcal{D}_{\text{kept}}}\!\!\sum_{t=1}^{|\mathbf{y}_{\text{teach}}|}\!\mathrm{KL}\!\left(\pi_{\mathrm{T}}^{(t)}\,\big\|\,\pi_\theta^{(t)}\right),
\label{eq:sfd}
\end{equation}
which is equivalent to the cross-entropy of $\pi_\theta$ under $\pi_{\mathrm{T}}$ on teacher-sampled tokens up to a $\theta$-independent entropy term. The KL is computed only over the positions of $\mathbf{y}_{\text{teach}}$. In same-size mode, $\pi_{\mathrm{T}} = \pi_{\theta_{\text{EMA}}}$; in cross-size mode, $\pi_{\mathrm{T}}$ is a frozen larger same-family model. Both modes use the same loss and the same training script and differ only in the teacher assignment.

\paragraph{When each mode wins.} The same-size mode requires no external model and no labels beyond the screenshot-click pairs already in $\mathcal{D}$, which makes it our recommended cold-start when serving a teacher alongside training would be expensive. The cross-size mode injects capability from a stronger model and is preferable when a same-family teacher is available cheaply.

\subsection{Reinforcement learning with GRPO}
\label{sec:rl}
We follow each cold-start with Group Relative Policy Optimization (GRPO). GRPO adapts the proximal-policy-optimization template to the verifiable-reward setting and removes the value network of standard PPO by computing advantages relative to a group of on-policy samples for the same prompt. We describe the algorithm in three steps.

\textbf{Step 1: rollout.} For each prompt $\mathbf{x}$ drawn from a training distribution $\mathcal{D}$, we sample a group of $G$ completions independently from the rollout policy,
\[\{\mathbf{o}^{(i)}\}_{i=1}^{G}\;\sim\;\pi_{\theta_{\text{old}}}(\cdot\mid\mathbf{x}).\]
Sampling uses a fixed temperature and a fixed maximum completion length. Each completion is scored by the verifiable reward
\begin{equation}
\begin{aligned}
r(\mathbf{o};\mathbf{x},\mathcal{B}_{\text{GT}}) =\;& 0.1\cdot\mathbf{1}[\mathbf{o}\text{ parses}] \\
&{}+ \exp\!\left(-\tfrac{1}{2}\!\left[\tfrac{(\Delta x)^2}{\sigma_x^2} + \tfrac{(\Delta y)^2}{\sigma_y^2}\right]\right),
\end{aligned}
\label{eq:reward}
\end{equation}
where $(\Delta x, \Delta y)$ is the offset from the predicted click to the ground-truth box center and $(\sigma_x, \sigma_y)$ are half the bounding-box width and height. All four quantities are in the absolute screen-pixel frame of the training screenshot, whose resolution $(W,H)$ varies per record because WinDOM draws viewports from a weighted pool (\cref{app:dataset}); the model emits integer coordinates on the $[0,1000]^2$ scale of \cref{sec:background}, which the reward function rescales by the per-record $(W,H)$ before computing $(\Delta x, \Delta y)$, so $\sigma_x$ and $\sigma_y$ always live in the same pixel frame as the bounding box. The first term encourages parsable output; the second term, with peak value one at the center, is a Gaussian-shaped click-in-target reward~\citep{Zhou2025GUIG1} that preserves variance in the group-relative advantage even when most rollouts land inside the bounding box.

\textbf{Step 2: group-normalized advantage.} GRPO replaces a learned value baseline with the empirical mean of the group's rewards. The advantage of the $i$-th completion is the standardized reward,
\begin{equation}
\hat{A}^{(i)}
\;=\; \frac{r^{(i)} - \mu(\{r^{(j)}\}_{j=1}^{G})}{\sigma(\{r^{(j)}\}_{j=1}^{G}) + \epsilon},
\label{eq:adv}
\end{equation}
where $\mu$ and $\sigma$ are the group mean and standard deviation and a small constant guards against zero variance. When all $G$ completions for a prompt receive the same reward, $\sigma\to 0$ and the advantage degenerates; the Gaussian-shaped reward in \cref{eq:reward} mitigates this by preserving variance across rollouts even when most of them land inside the bounding box.

\textbf{Step 3: clipped surrogate update.} Let $\rho^{(i)}(\theta)=\pi_\theta(\mathbf{o}^{(i)}\mid\mathbf{x})/\pi_{\theta_{\text{old}}}(\mathbf{o}^{(i)}\mid\mathbf{x})$ denote the importance ratio. We maximize
\begin{equation}
\begin{aligned}
\mathcal{J}_{\text{GRPO}}(\theta) = \mathbb{E}\Big[&\tfrac{1}{G}\!\sum_{i=1}^{G}\!\min\!\bigl(\rho^{(i)}\hat{A}^{(i)},\\
&\mathrm{clip}(\rho^{(i)}, 1{-}\varepsilon, 1{+}\varepsilon)\,\hat{A}^{(i)}\bigr)\Big]
\end{aligned}
\end{equation}
with a fixed clip range $\varepsilon=0.2$ (\cref{app:hp}). We do not add a Kullback-Leibler regularizer toward the cold-start policy ($\beta=0$); the cold-start enters only as the initialization of $\theta$.

The choice of group size and sampling temperature trades off rollout cost against the variance of the empirical baseline in \cref{eq:adv}; concrete values are reported with the experiments in \cref{sec:experiments}.

\subsection{Two RL regimes}
\label{sec:two-regimes}
The central experimental contrast of the paper compares two ways of seeding the GRPO stage. \emph{Late-init RL} starts from a saturated cold-start, that is, a checkpoint near the end of supervised training. \emph{Early-init RL} starts from an early cold-start, before supervised training has converged. Both regimes share their reinforcement learning hyperparameters; only the initialization differs. We instantiate both regimes for all three cold-start regimes, yielding six reinforcement learning configurations.

\subsection{Evaluation protocol}
\label{sec:eval}
All evaluations use greedy decoding, verified bit-deterministic across reruns in \cref{sec:variance}. We report click-in-target accuracy on the in-distribution held-out WinDOM split, a deterministic $2{,}000$-example subsample of the full $6{,}070$-record test set fixed across all comparisons, and on three out-of-distribution real-screenshot benchmarks: ScreenSpot-Pro on professional desktop applications, OSWorld-G on open-domain operating-system tasks, and ScreenSpot-V2 on a broad mobile, web, and desktop mix, with respective sizes $1{,}530$, $510$, and $1{,}272$. The unweighted mean of the three OOD scores is reported as \emph{OOD-mean}.

\section{Experiments}
\label{sec:experiments}

Throughout, the base model we train is \textbf{Qwen3.5-2B}, the $\sim$2B-parameter member of the Qwen3.5 vision-language family; every trained row in this paper (Direct SFT, both SFD modes, and all GRPO runs) starts from this same 2B base. The larger $\sim$4B same-family checkpoint appears \emph{only} as an optional frozen teacher for the cross-size SFD mode (\cref{sec:sfd}); it is never trained, never the deployed policy, and is not required by the recommended same-size EMA recipe. The full experimental setup, including backbone choice, tool-call schema and parser, training schedule, GPU choice, and evaluation decoding parameters, is deferred to \cref{app:setup,app:hp,app:compute}. \Cref{sec:main-results} reports the main results, \cref{sec:findings} groups our three empirical findings, and \cref{sec:variance} reports a sanity check on variance.

\subsection{Main results}
\label{sec:main-results}

\begin{table}[t]
\centering
\small
\caption{Click-in-target accuracy (\%). Each cold-start is followed by two GRPO variants: \emph{+RL Early} initialises GRPO from step $100$ of the cold-start, \emph{+RL Late} from step $900$. We report each row's best-OOD-mean checkpoint; OOD is the unweighted mean over SS-Pro, OSG, and SS-V2. \textbf{Bold} = best, \underline{underline} = second among trained rows. Qwen3.5-4B is the cross-size SFD teacher, not trained here and not evaluated on SS-V2.}
\label{tab:main}
\renewcommand{\arraystretch}{1.10}
\setlength{\tabcolsep}{4pt}
\begin{tabular}{l c c c c c}
\toprule
Model & In-dist. & SS-Pro & OSG & SS-V2 & OOD \\
\midrule
Qwen3.5-2B (base)                  & 70.0 & 52.2 & 47.5 & 83.0 & 60.9 \\
Qwen3.5-4B (base)                  & 71.2 & 60.0 & 56.3 & 93.7 & 70.0 \\
\midrule
SFT                                & \textbf{82.1} & 47.6 & 49.0 & 85.9 & 60.8 \\
\rowcolor{gray!10}\quad +RL Early  & 77.2 & 52.7 & \underline{52.4} & 87.7 & 64.3 \\
\quad +RL Late                     & \underline{80.8} & 50.0 & 51.6 & 87.6 & 63.1 \\
\midrule
SFD-EMA                  & 76.8 & 53.5 & 51.2 & 88.8 & 64.5 \\
\rowcolor{gray!10}\quad +RL Early  & 75.4 & \underline{55.1} & 51.6 & \underline{89.0} & \underline{65.2} \\
\quad +RL Late                     & 79.0 & 53.2 & 51.0 & 85.4 & 63.2 \\
\midrule
SFD-4B                   & 76.5 & 54.7 & 50.8 & \textbf{89.5} & 65.0 \\
\rowcolor{gray!10}\quad \textbf{+RL Early} & 76.5 & \textbf{55.7} & \textbf{54.5} & 88.8 & \textbf{66.3} \\
\quad +RL Late                     & 79.7 & 53.0 & 51.4 & 88.9 & 64.4 \\
\bottomrule
\end{tabular}
\end{table}

\begin{figure}[t]
\centering
\includegraphics[width=\columnwidth]{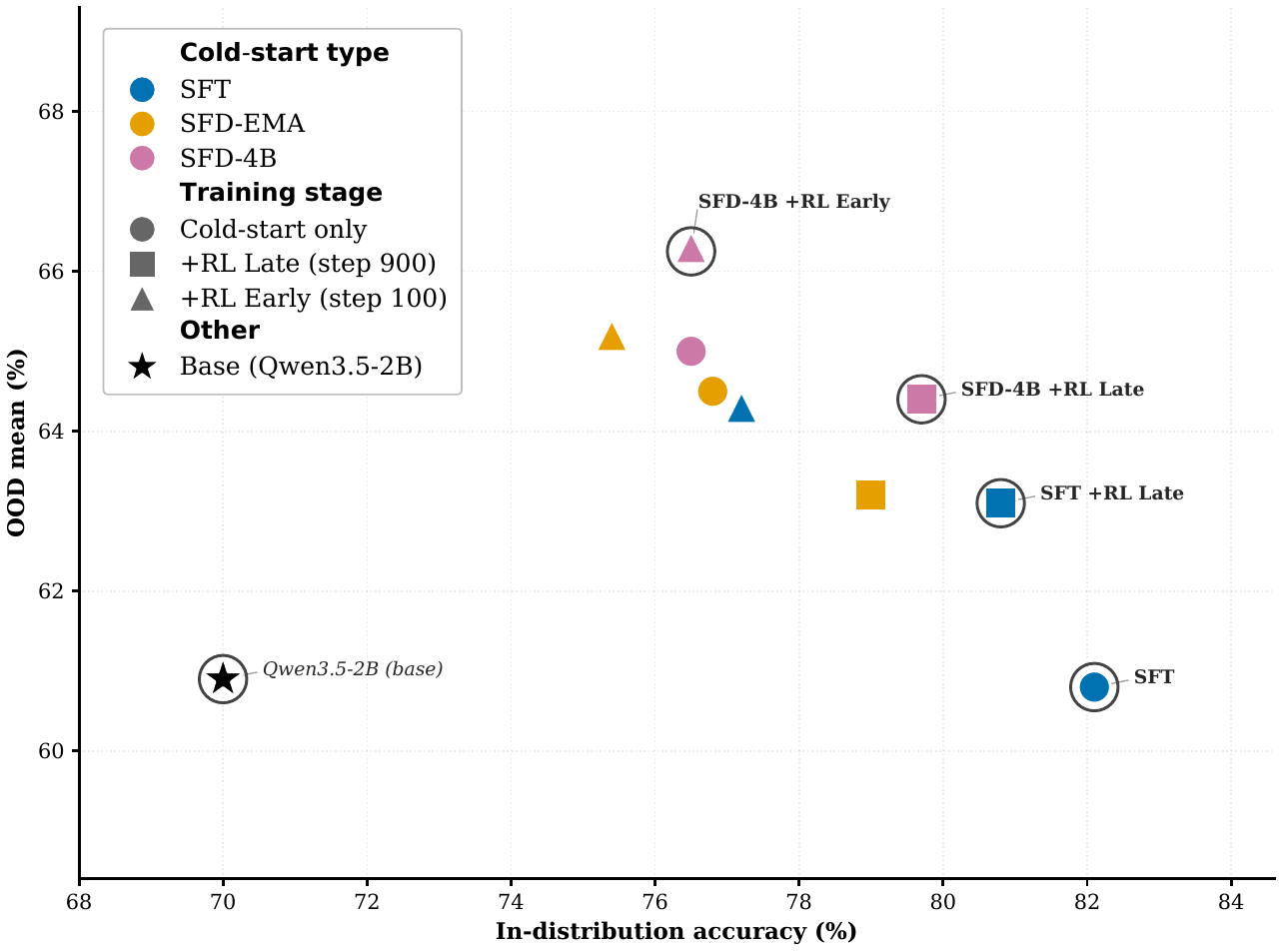}
\caption{Best-checkpoint accuracy of every training run: in-distribution vs.\ OOD-mean (the three-benchmark mean of \cref{tab:main}). Shape encodes stage (circle: cold-start, square: +RL Late, triangle: +RL Early); color encodes cold-start type (blue: SFT, orange: SFD-EMA, magenta: SFD-4B). The supervised cold-start, the late-init RL run, and our best configuration are highlighted.}
\label{fig:pareto}
\end{figure}

\Cref{tab:main} reports the best checkpoint per model on each benchmark; the Qwen3.5-2B and 4B bases are provided as references. Early-init RL on top of the cross-size SFD cold-start is the strongest of our nine configurations on the three-benchmark OOD mean: it leads on ScreenSpot-Pro and OSWorld-G among the trained rows, stays within one point of the strongest in-distribution row on ScreenSpot-V2, and over the Qwen3.5-2B base it gains $+3.5$ pp on ScreenSpot-Pro ($55.7$ vs $52.2$), $+7.0$ pp on OSWorld-G ($54.5$ vs $47.5$), $+5.8$ pp on ScreenSpot-V2 ($88.8$ vs $83.0$), and $+5.4$ pp on the three-benchmark OOD-mean ($66.3$ vs $60.9$). \Cref{fig:pareto} visualizes the same data as a scatter plot in the in-distribution / out-of-distribution plane, with marker shape encoding the training stage (circle: cold-start only, square: +RL Late, triangle: +RL Early) and marker colour encoding the cold-start type (blue: SFT, orange: SFD-EMA, magenta: SFD-4B): the Early-init RL family sits on the upper-left frontier, while the Late-init RL family clings to the cold-start curve.

\subsection{Findings}
\label{sec:findings}

\begin{figure}[t]
\centering
\includegraphics[width=\columnwidth]{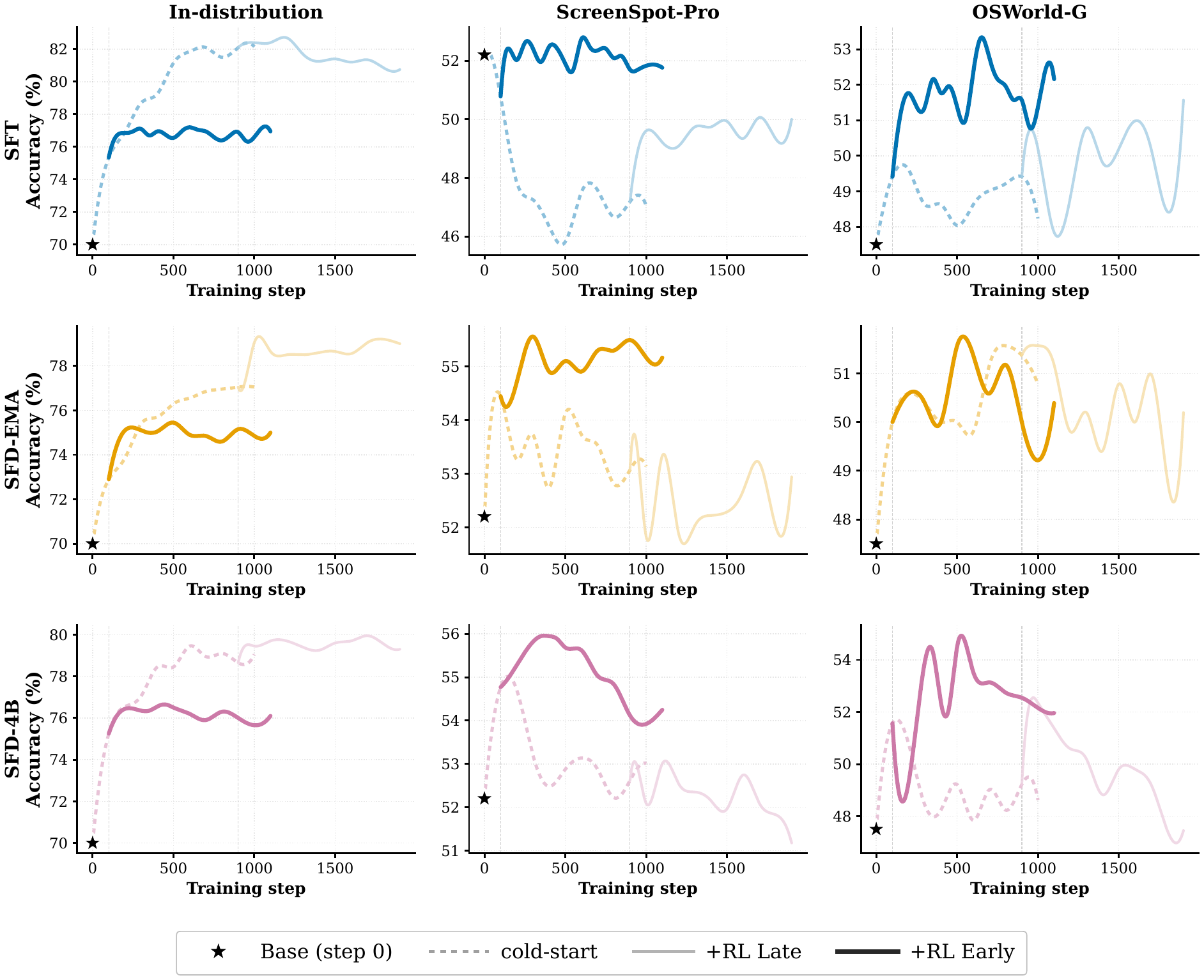}
\caption{Click-in-target accuracy vs.\ training step. Rows: cold-start method (SFT, SFD-EMA, SFD-4B); columns: benchmark (in-distribution, ScreenSpot-Pro, OSWorld-G). Each cell shows the cold-start curve (gray dotted), the \emph{+RL Late} path (gray solid, cold-start to step 900 then 1000 GRPO steps), and our best path \emph{+RL Early} (accent colour, cold-start to step 100 then 1000 GRPO steps); vertical dashed lines mark the two GRPO start checkpoints. The SFT trap is visible in the top-row, middle-column panel (SFT $\times$ ScreenSpot-Pro): the SFT cold-start drops $\sim$4 points and the \emph{+RL Late} path inherits that low operating point, whereas the SFD rows rise or stay flat and \emph{+RL Early} climbs further. ScreenSpot-V2 trajectories are qualitatively similar and reported at each row's best checkpoint in \cref{tab:main}.}
\label{fig:sft-trap}
\end{figure}

\paragraph{F1: Supervised fine-tuning trades OOD accuracy for in-distribution.}
\label{sec:finding1}
Direct SFT exhibits a clean memorization signature (\cref{fig:sft-trap}, top row): in-distribution accuracy rises from the Qwen3.5-2B base value of $70.0\%$ to $82.1\%$ over $1{,}000$ steps ($+12.1$pp), while ScreenSpot-Pro accuracy decays from the base value of $52.2\%$ down to $45.8\%$ at step $500$ and recovers only partially to $47.1\%$ by step $1{,}000$ ($-5.1$pp net), with the worst ScreenSpot-Pro checkpoint sitting near the conventional one-epoch stopping point. The two effects almost exactly cancel on the OOD-mean axis: a fully converged SFT cold-start ends within a tenth of a point of the Qwen3.5-2B base ($60.8$ vs $60.9$ in \cref{tab:main}). Both Self-Family Distillation cold-starts soften the trade: ScreenSpot-Pro stays roughly flat across the cold-start phase rather than decaying, and the supervised stage already lifts the OOD-mean to $64.5$ (SFD-EMA) and $65.0$ (SFD-4B) before reinforcement learning begins.

\paragraph{F2: Cold-start depth controls RL transfer.}
\label{sec:finding2}
The same GRPO recipe yields opposite OOD outcomes under the two cold-start depths. Late-init RL holds the in-distribution operating point in a narrow $79$--$81\%$ band and barely moves ScreenSpot-Pro, but its OSWorld-G trajectory drifts downward: on the SFD-4B run, OSG enters GRPO at the cold-start value of $49.2$, briefly rises to $52.4$ at step $100$, and then decays back to $47.5$ at step $1{,}000$, the same value as the Qwen3.5-2B base, undoing both the cold-start gain and the brief in-RL bump.

Early-init RL takes the opposite trade. It gives up roughly $3$--$4$ points of in-distribution accuracy and converts that headroom into clear OOD progress: on the same SFD-4B family, the Early-init trajectory peaks per-checkpoint at $55.9$ on ScreenSpot-Pro and $54.5$ on OSWorld-G (the headline row in \cref{tab:main} reports $55.7$ and $54.5$ at the best-OOD-mean checkpoint), and these peaks sit above \emph{every} matched Late-init checkpoint we evaluate. The dominance is family-wide, not just at the peak: Early-init beats Late-init on three-benchmark OOD-mean at every GRPO step, by margins that grow consistently with cold-start quality: $63.1\!\to\!64.3$ for SFT ($+1.2$), $63.2\!\to\!65.2$ for SFD-EMA ($+2.0$), and $64.4\!\to\!66.3$ for SFD-4B ($+1.9$) (\cref{fig:sft-trap}).

The cold-start ledger pushes the same way. Late-init enters GRPO from a much longer cold-start ($900$ SFT steps vs Early-init's $100$) and runs the same $1{,}000$-step GRPO phase on top, yet still loses on every OOD axis. Early-init therefore wins from a \emph{less-trained} initialisation, a result that runs against the natural intuition that a more thoroughly converged cold-start should make it easier for GRPO to generalise out of distribution.
\begin{remark}[Hypothesis: entropy collapse]
\label{rem:entropy-collapse}
A natural account of why late-init RL fails to lift the OOD operating point is entropy collapse. A saturated cold-start has been driven by cross-entropy onto a near-deterministic action distribution: the per-prompt rollout group sampled from $\pi_\theta$ is concentrated on a single coordinate, so the within-group standard deviation $\sigma$ in \cref{eq:adv} approaches zero and the group-relative advantage degenerates. Under that condition GRPO has no signal with which to refine OOD behaviors, since every rollout receives the same reward and contributes a zero-advantage update. The same pathology has been documented for clipped policy gradients in mathematical reasoning by \citet{Yu2024DAPO}, and the broader observation that heavy supervised pre-training collapses the policy entropy that subsequent RL relies on has been reported by \citet{DeepSeek2025R1,Chu2025SFTMemRL,Shao2024DeepSeekMath}. We have observed this signature in our training logs: the saturated SFD-4B cold-start enters GRPO with markedly lower per-token sampling entropy and a higher fraction of zero-variance rollout groups than the early cold-start, and the gap persists throughout the GRPO run. The empirical OOD contrast above is consistent with this account, though we do not claim it isolates entropy from alternative mechanisms (effective-update budget, on-policyness, distance from base); a quantitative entropy/variance plot across saturation levels is left to future work.
\end{remark}

\paragraph{F3: Self-Family Distillation is robust to teacher choice.}
\label{sec:finding3}
Both SFD modes track each other closely under Early-init RL: across every evaluated GRPO checkpoint, in-distribution accuracy stays in the $74.6$ to $76.5\%$ band and OOD-mean within $\pm 1.5$ percentage points (\cref{fig:sft-trap}). The same-size EMA mode trails the cross-size $4$B mode by roughly one OOD-mean point overall ($65.2$ vs $66.3$): the two are essentially tied on ScreenSpot-V2 ($89.0$ vs $88.8$), within $0.6$ points on ScreenSpot-Pro ($55.1$ vs $55.7$), and separated by $2.9$ points on OSWorld-G ($51.6$ vs $54.5$), where the residual gap concentrates. The practical takeaway is that the same-size mode, which needs no external model, is the recommended default; the cross-size mode is worth its extra compute only when a stronger same-family teacher is available cheaply. The two modes also differ in how often the teacher's rollout passes the click-in-bounding-box filter. With batch size $2$, each training micro-batch contains two teacher rollouts; the micro-batch is kept and contributes a gradient if at least one of those two rollouts lands inside the GT bounding box, otherwise the entire micro-batch is discarded. Under that definition, $81.2\%$ of micro-batches contribute gradients in the cross-size $4$B mode and $62.1\%$ in the same-size EMA mode (so $18.8\%$ and $37.9\%$ of micro-batches respectively are dropped at training time). The lower EMA kept-rate is consistent with the EMA teacher's lower per-step competence and confirms that the rejection-sampling filter remains a non-trivial verifier in either mode.

\subsection{Variance}
\label{sec:variance}

\paragraph{Eval determinism.} Greedy decoding makes evaluation deterministic up to numerical noise. We reran the SFD-4B Early-init RL checkpoint at GRPO step $400$ (the row that bolds OSWorld-G in \cref{tab:main}) three times on different H100 GPUs. All three runs returned exactly $278/510=54.51\%$ on OSWorld-G with identical per-category breakdowns, and exactly $852/1530=55.69\%$ on ScreenSpot-Pro. The only source of variance in our reported numbers is therefore training itself, not evaluation.

\paragraph{Training variance.} To bound the training-side noise of our headline result, we trained the SFD-4B Early-init RL configuration under three independent random seeds drawn at the trainer level, holding every other element of the recipe fixed: the cold-start checkpoint that initialises GRPO is byte-identical across the three runs, evaluation is greedy and therefore deterministic, and only the optimiser, data sampler, and rollout RNGs differ. The exact seed-propagation path through our training stack is described in \cref{app:variance}. On the best GRPO model the cross-seed accuracy range is $0.9$pp on the in-distribution split, $0.6$pp on ScreenSpot-Pro, and $0.4$pp on OSWorld-G. The cross-seed spread is therefore smaller than the gap between SFD-4B Early-init RL and any Late-init RL or cold-start-only configuration in \cref{tab:main} on every benchmark we report, and we treat the headline numbers as effectively reproducible across seeds.

\section{Discussion}
\label{sec:discussion}

We close by relating the recipe to safety properties for computer-use agents, then noting where the conclusions of \cref{sec:experiments} are most fragile and where the recipe is most likely to extend.

\subsection{Provenance of the supervision channel}
\label{sec:safety}
Every supervisory signal in the recipe, the rejection sampler at the SFD cold-start and the GRPO reward (\cref{eq:reward}), reduces to a single deterministic point-in-bounding-box check; no learned reward model, preference-data ranker, or LLM judge enters the loop. The cold-start labels are read from the renderer's own layout rather than from OCR or human raters, eliminating a class of label-noise failures (mislabelled crops, instruction--bbox swaps) at source. The resulting property is narrow: an outside reviewer can re-derive both the labels and the verifier from the released regeneration script, without trusting our compute environment. We make no corresponding claim about the behaviour of the trained policy at deployment.

\subsection{Limitations}
\label{sec:limitations}
Our findings should be read against three scope restrictions.\\[4pt]
\textbf{Single backbone size:} we evaluate one $2$B backbone, leaving open whether the early-cold-start advantage transfers to the $7$B and $72$B regimes.\\[4pt]
\textbf{Single model family:} both the student and the cross-size SFD teacher come from the Qwen3.5 family; we have not tested whether the same recipe works across families with different tokenisers or coordinate-emission protocols.\\[4pt]
\textbf{Static evaluation:} our evaluation is single-click and static rather than agentic; we report transfer to real-screenshot grounding benchmarks but not to multi-step closed-loop environments.

\subsection{Future work}
Several directions follow naturally from the recipe and the artefact.

\paragraph{Online reinforcement learning.} Because our pipeline includes both the dataset generator and the underlying renderer, WinDOM doubles as a fully autonomous environment: the same Win11React process that produces the static training corpus can serve on-policy rollouts at training time. The natural next step is to take GRPO online, using the renderer as a closed-loop reinforcement-learning environment that streams fresh, verifier-checked rollouts.

\paragraph{Online self-distillation.} The same shift applies to the SFD cold-start: instead of pre-collecting a rejection-sampled corpus, an EMA self-teacher can generate supervision on the fly, with the click-in-bounding-box filter applied per-step.

\paragraph{Negative-rollout signal.} The current rejection sampler discards every teacher rollout that misses the GT bounding box; with the deterministic verifier already in place, those rejected rollouts are a natural source of \emph{negative} supervision for contrastive losses or preference-optimisation objectives such as DPO. Folding negatives into the cold-start would turn what is currently a wasted budget into a cheap form of preference data.

\paragraph{Long-horizon planning.} Beyond single-click grounding, the same environment can host multi-step UI workflows that span several screens, opening a path from a single-click grounding agent to a planning agent within the same WinDOM training loop.

\paragraph{Verifier-only post-training beyond GUI grounding.} The recipe (deterministic verifier, filtered same-family distillation, under-saturated cold-start) applies to any task that admits a programmatic ground-truth check, including refusal-format compliance, JSON-schema validity, code-execution unit tests, and math-answer equality; whether the cold-start-depth lever transfers to those tasks at 2B scale is an open empirical question.

\paragraph{Adversarial-UI evaluation.} Scripting the same renderer with decoy DOM nodes, prompt-injected button labels, or spoofed accessibility roles converts Win11React from a training corpus into a closed-loop probe, which is the natural follow-up to the training-time-only safety claims of \cref{sec:safety-impact}.

\subsection{Broader Impact and Safety}
\label{sec:safety-impact}
\paragraph{Dual use.} Reaching $55.7$ / $54.5$ / $88.8$ on ScreenSpot-Pro, OSWorld-G, and ScreenSpot-V2 at 2B parameters and zero human bounding-box supervision lowers the absolute cost of building UI-driving agents, for both legitimate use (assistive desktop control, accessibility replacements, automated regression testing) and unattended abuse (credential stuffing, ad-fraud clickers). Withholding a 2B checkpoint does not meaningfully change this threat model relative to existing open GUI agents; the artefact reduces cost, not ceiling.

\paragraph{Auditability.} Because WinDOM is regenerable from Win11React together with the released collection script, every $(\mathbf{x},\mathcal{B}_{\text{GT}})$ pair is reconstructible by a third party, and the geometric verifier that scores it is deterministic. \Cref{sec:variance} additionally verifies bit-determinism of greedy evaluation across H100 GPUs on both OSWorld-G ($278/510$) and ScreenSpot-Pro ($852/1530$), so any reported accuracy can be independently re-derived from the released checkpoint. These are properties of the training and evaluation pipeline, not of the deployed policy.

\paragraph{Scope of the safety claim.} What the design buys is supervision-channel auditability. We do not claim that the trained policy is safe to deploy unsupervised, robust to adversarial UI or prompt injection, or that the click-in-bounding-box reward functions as an alignment objective; an agent that clicks the right pixel can still be steered into harmful actions by the calling policy. Verifying any of those properties requires the dedicated experiments (adversarial-instruction evaluation, decoy-DOM probes, red-teaming inside the renderer) that we leave to future work.

\section{Conclusion}

Three takeaways follow from the experiments, and they together aim at one goal: making small grounding models work well on out-of-distribution screenshots. The browser DOM is a free, dense, layout-rectangle source of grounding labels, so any open-source web reimplementation of an operating system is a high-fidelity training corpus at zero annotation cost. In its same-size EMA mode, Self-Family Distillation provides a teacher-free cold-start that approaches the cross-size 4B-teacher variant within roughly one OOD-mean point ($65.2$ vs $66.3$, \cref{tab:main}) and lets small students avoid the cost of running a larger model alongside training. And cold-start depth is a hyperparameter that small models cannot ignore: the same RL recipe yields opposite OOD outcomes depending on whether the supervised base has been allowed to converge. Together they let a 2B vision-language model trained on synthetic Windows~11 data reach $88.8\%$ on ScreenSpot-V2 and $55.7\%$ on ScreenSpot-Pro without a single human bounding-box annotation.

\section*{Reproducibility Statement}
We will release the code, training and evaluation scripts, the WinDOM corpus, the rendering pipeline, and the trained model weights under a permissive license at the camera-ready stage. All hyperparameters, system prompts, compute budgets, dataset details, full per-checkpoint accuracies, evaluation-determinism evidence, three-seed replication, and the headline-row reproduction recipe are reported in the appendix.

\paragraph{Upstream license and trade-dress.} WinDOM is rendered against \texttt{blueedgetechno/win11React}, a community web reimplementation of Windows~11 released under the CC0-1.0 license, which permits redistribution of derivative renderings. The visual elements that the Win11React project itself replicates (window-control glyphs, the Start menu chrome, application icons, default Windows-style wallpapers) reproduce Microsoft trade-dress and are not licensed by Microsoft. To avoid redistributing trade-dress under our name we will, at camera-ready, (i) ship WinDOM as DOM snapshots, bounding boxes, and instructions plus a deterministic regeneration script that re-renders screenshots locally from Win11React rather than as a binary screenshot dump, and (ii) replace the curated wallpaper bundle with a permissively-licensed (CC0 / CC-BY) wallpaper set so that none of the redistributed image bytes originate from Microsoft assets. The exact license under which the regeneration script and the (instruction, bounding-box, DOM) tuples are released will be confirmed before camera-ready, and we will document any user-side caveats around running the regeneration pipeline on third-party trade-dress.

\section*{Broader Impact}
Improved GUI grounding supports accessibility tools, automated software testing, and assistive desktop agents. The same capability enables misuse such as captcha bypass, scraping, and unauthorized account access. Our pipeline trains on a synthetic clone of Windows~11 and contains no personal data. Downstream agentic systems should add rate limiting, sandboxing, and human confirmation for destructive actions.

\bibliography{main}
\bibliographystyle{icml2026}

\newpage
\appendix
\twocolumn
\linespread{1.05}\selectfont
\setlength{\parskip}{2pt}
\setlength{\abovecaptionskip}{4pt}
\setlength{\belowcaptionskip}{2pt}
\setlength{\textfloatsep}{6pt plus 1pt minus 1pt}
\setlength{\intextsep}{6pt plus 1pt minus 1pt}

\section{Notation summary}
\label{app:notation}

This section consolidates every symbol, acronym, and named regime used in the paper into a single one-line glossary, so that the appendix can be read end to end without backward navigation.

\paragraph{Symbols.}
\begin{itemize}\itemsep1pt
\item $\pi_\theta(\mathbf{o}\mid\mathbf{x})$: the student policy, a vision-language transformer that autoregressively samples an output sequence $\mathbf{o}$ given input $\mathbf{x}$ (a screenshot plus a textual instruction). See \cref{sec:model-class}.
\item $\pi_{\mathrm{T}}$: the teacher policy. In same-size SFD, $\pi_{\mathrm{T}}=\pi_{\theta_{\text{EMA}}}$; in cross-size SFD, $\pi_{\mathrm{T}}$ is a frozen larger same-family checkpoint.
\item $\pi_{\mathrm{S}}$: the student policy under training. Used interchangeably with $\pi_\theta$ in figures.
\item $\theta$: the trainable parameters of the student.
\item $\theta_{\text{EMA}}$: the exponential moving average of the student parameters with decay $\alpha$, updated as $\theta_{\text{EMA}}\!\leftarrow\!\alpha\,\theta_{\text{EMA}} + (1-\alpha)\,\theta$ on every optimizer step.
\item $\alpha$: EMA decay (we use $\alpha=0.99$).
\item $\mathbf{x}$: input to the policy (system prompt, screenshot, user instruction).
\item $\mathbf{o}=(o_1,\ldots,o_T)$: output token sequence over the vocabulary $\mathcal{V}$.
\item $\mathbf{y}\subseteq\mathbf{o}$: the structured action sub-sequence whose tokens encode the predicted click coordinate.
\item $\mathbf{y}_{\text{teach}},\,\mathbf{y}_{\text{student}}$: teacher and student action sub-sequences during distillation.
\item $\mathrm{decode}(\cdot)$: a fixed parser $\mathcal{V}^*\!\to\!([0,W]\!\times\![0,H])\cup\{\bot\}$ that returns either a click point in pixels or $\bot$ for non-parsable rollouts. $\bot$ counts as an incorrect click.
\item $\mathcal{B}_{\text{GT}}$: ground-truth bounding box (axis-aligned rectangle) on the screenshot.
\item $(W,H)$: pixel dimensions of the screenshot. The model emits coordinates on a normalized $[0,1000]^2$ grid that the parser rescales to $(W,H)$.
\item $G$: rollout group size in GRPO (we use $G=8$).
\item $T_{\text{rollout}}$: rollout sampling temperature (we use $1.3$).
\item $r(\mathbf{o};\mathbf{x},\mathcal{B}_{\text{GT}})$: verifiable reward, the sum of a $0.1$ format bonus and a Gaussian click-in-target term (\cref{eq:reward}).
\item $(\Delta x, \Delta y)$: pixel offset from predicted click to ground-truth box center.
\item $(\sigma_x, \sigma_y)$: half the GT-box width and height in pixels; sets the Gaussian reward bandwidth.
\item $\hat{A}^{(i)}$: group-normalized advantage of the $i$-th rollout (\cref{eq:adv}).
\item $\rho^{(i)}(\theta)$: importance ratio $\pi_\theta(\mathbf{o}^{(i)}\!\mid\!\mathbf{x})/\pi_{\theta_{\text{old}}}(\mathbf{o}^{(i)}\!\mid\!\mathbf{x})$.
\item $\varepsilon$: PPO-style clip range on $\rho^{(i)}$ (we use $\varepsilon=0.2$).
\item $\beta$: KL coefficient toward the cold-start policy in GRPO (we use $\beta=0$).
\item $\mathcal{L}_{\text{SFD}}$: forward-KL distillation loss at teacher-token positions (\cref{eq:sfd}).
\item $\mathcal{J}_{\text{GRPO}}$: clipped surrogate objective maximized in the RL stage.
\item $\mathcal{D}$: training distribution over (screenshot, instruction, GT-box) triples; $\mathcal{D}_{\text{kept}}$ is the subset on which the teacher rollout passed the click-in-bbox filter.
\end{itemize}

\paragraph{Acronyms and named regimes.}
\begin{itemize}\itemsep1pt
\item GUI: graphical user interface.
\item DOM: document object model. The browser's tree of layout boxes; we read interactable bounding boxes from \texttt{getBoundingClientRect} on each DOM node.
\item ARIA: accessible rich internet applications, the W3C accessibility role vocabulary used to filter interactable elements.
\item SFT: supervised fine-tuning on (screenshot, instruction, action) tuples with the cross-entropy loss.
\item SFD: Self-Family Distillation, our umbrella for two cold-start variants whose teacher is from the student's own model family.
\item SFD-EMA: same-size SFD; the teacher is an EMA of the student.
\item SFD-4B: cross-size SFD; the teacher is a frozen larger same-family checkpoint (in our instantiation a 4B parameter sibling of the 2B student).
\item GRPO: Group Relative Policy Optimization; the RL algorithm of \cref{sec:rl}.
\item KL: Kullback-Leibler divergence, used both as the SFD distillation loss and as an optional GRPO regularizer.
\item Cold-start: the supervised checkpoint that initializes RL.
\item Saturated cold-start: a cold-start whose in-distribution training-accuracy curve has plateaued (we report step $900$ of $1{,}000$).
\item Early cold-start: a pre-plateau cold-start (we report step $100$ of $1{,}000$).
\item Late-init RL: GRPO launched from a saturated cold-start.
\item Early-init RL: GRPO launched from an early cold-start. The SFD-4B Early-init RL row is our headline configuration.
\item In-distribution (in-dist.): a fixed $2{,}000$-record subsample of the held-out WinDOM test split.
\item OOD: out-of-distribution. We report three external benchmarks (ScreenSpot-Pro, OSWorld-G, ScreenSpot-V2) and their unweighted mean (OOD-mean).
\item SS-Pro, OSG, SS-V2: shorthand column heads for ScreenSpot-Pro, OSWorld-G, and ScreenSpot-V2 in \cref{tab:main}.
\item Click-in-target accuracy: the fraction of evaluation instances where the predicted click point lies inside $\mathcal{B}_{\text{GT}}$ (\cref{eq:cgt}). The only metric reported in this paper.
\item WinDOM: our DOM-grounded dataset built from a community web reimplementation of Windows~11.
\item QA-pass: the verification stage where a second-pass model classifies each candidate as keep, edit, or reject (\cref{sec:dataset}).
\item bbox: bounding box.
\item pp: percentage points.
\end{itemize}

\section{Experimental setup}
\label{app:setup}

This section gives the concrete instantiation and the full training schedule that lie behind every reported number. Together with the hyperparameters in \cref{app:hp}, the dataset construction in \cref{app:dataset}, and the prompt schema in \cref{app:prompts}, it suffices to reproduce a single training run from scratch on equivalent hardware.

\paragraph{Backbone choice and rationale.} The base model trained in all experiments is \textbf{Qwen3.5-2B}, the $\sim$2B-parameter vision-language model in the Qwen3.5 family \citep{Qwen2025Qwen3VL}; every trained checkpoint we report (Direct SFT, SFD-EMA, SFD-4B, and all GRPO runs) is this same 2B base, and the $\sim$4B same-family checkpoint is used only as an optional frozen teacher in the cross-size SFD mode, never as the trained or deployed model. We picked this family for three concrete reasons. First, it ships with a native tool-call interface (a JSON action block delimited by \texttt{<tool\_call>}\ldots\texttt{</tool\_call>}) whose left-click action takes a coordinate on a normalized $[0,1000]^2$ grid; this lets us reuse the model's pretrained tool-call vocabulary without adding an auxiliary grounding head, vocabulary expansion, or any custom modeling code. Second, a $\sim$4B-parameter same-family sibling shares the tokenizer and special-token table with the 2B student, which is exactly the alignment that the cross-size SFD loss \cref{eq:sfd} requires (token-level forward KL only makes sense over a shared vocabulary). Third, both the 2B student and the 4B teacher are publicly available, run in standard \texttt{transformers} on a single H100 or H200 GPU, and produce the structured JSON actions that the parser below consumes. Replacing the backbone with a same-family pair from a different lineage (for example a $\sim$2B and a $\sim$4B from any other vision-language family with matching tokenizers and a documented coordinate emission convention) requires no change to the training recipe other than swapping the system prompt template (\cref{app:prompts}).

\paragraph{Coordinate emission and parsing.} At both training and evaluation time, the model is asked to emit a single tool call describing a left click. Concretely, the structured action is a JSON object with a tool name (\texttt{computer\_use}) and an \texttt{arguments} field containing a \texttt{coordinate} entry. The \texttt{coordinate} entry is either a 2-tuple $(x, y)$ in $[0,1000]^2$ (a click point) or a 4-tuple $(x_1, y_1, x_2, y_2)$ (a bounding-box prediction; we reduce it to its center for scoring). The parser (i) extracts everything between the first \texttt{<tool\_call>} and the matching \texttt{</tool\_call>}, (ii) parses it as JSON, (iii) reads \texttt{arguments.coordinate}, (iv) reduces a 4-tuple to its center, and (v) rescales the resulting integer point from the $[0,1000]^2$ grid to absolute pixel coordinates by multiplying by $(W/1000, H/1000)$ where $(W,H)$ is the screenshot resolution at evaluation time. Anything that fails to parse, or whose structured action is missing or malformed, returns $\bot$ and counts as an incorrect click. The same parser is shared by training-time reward computation (\cref{eq:reward}), evaluation-time accuracy (\cref{eq:cgt}), and the click-in-bbox filter that gates teacher rollouts in SFD.

\paragraph{Decoding settings.} All evaluations use greedy decoding: the model takes its argmax token at every position. We disable sampling, set the temperature to $1.0$ (which is inert under argmax decoding), and disable top-$k$ and top-$p$ truncation. This makes the decoder bit-deterministic on a fixed GPU; small numerical differences across GPU types are discussed in \cref{app:determinism}. The maximum new-token budget is $128$ for the in-distribution split, OSWorld-G, and ScreenSpot-V2, and $1024$ for ScreenSpot-Pro because the Pro instructions are noticeably longer and the model occasionally produces a longer chain of tool-internal text before emitting its action. At these budgets the parse-failure rate is below $0.1\%$ on every benchmark we evaluate; we did not observe a benchmark-by-benchmark sensitivity to the exact budget within $\pm 50\%$ of the values above.

\paragraph{Training schedule.} The supervised cold-start runs for $1{,}000$ optimizer steps in all three regimes (Direct SFT, SFD-EMA, SFD-4B), saving a checkpoint every $100$ steps. We refer to the step-$900$ checkpoint as the \emph{saturated} cold-start (its in-distribution training-accuracy curve has plateaued by then) and the step-$100$ checkpoint as the \emph{early} cold-start. Both SFD modes present their teacher with the ground-truth bounding box jittered by $\pm 30$ on the $[0,1000]^2$ scale (independent integer noise per coordinate, clamped to $[0,1000]$); the student is never shown the hint, only the teacher is. The same-size SFD-EMA mode uses parameter decay $\alpha=0.99$, that is, $\theta_{\text{EMA}}\!\leftarrow\!0.99\,\theta_{\text{EMA}} + 0.01\,\theta$ applied on every optimizer step. GRPO runs for an additional $1{,}000$ steps starting from the chosen cold-start checkpoint, again saving every $100$ steps; we report the GRPO checkpoint with the best mean accuracy across the three OOD benchmarks. Hardware: cold-start uses a single H200 GPU; GRPO and evaluation use a single H100 GPU. Wall-clock and total budgets are in \cref{app:compute}; the optimizer and reward hyperparameters are in \cref{app:hp}.

\section{Hyperparameters}
\label{app:hp}

\Cref{tab:hp-cold,tab:hp-grpo} list every numeric hyperparameter used in cold-start and GRPO training. The same values were used for all rows of \cref{tab:main}, with the single exception that the cross-size SFD teacher is absent in Direct SFT and SFD-EMA. Anything not listed (gradient clipping, weight decay, dropout) is left at the framework default for the underlying training library; we do not introduce dropout, weight tying, or auxiliary heads beyond the backbone's own.

\paragraph{Cold-start hyperparameter notes.} The learning rate ($5\times 10^{-6}$) is at the low end of the typical SFT range for $\sim$2B vision-language models and was chosen to keep the SFD-EMA teacher (a moving average of the student) close to the student during the first epoch; raising the learning rate by an order of magnitude makes the EMA teacher chase a moving target and degrades F3 in our preliminary runs. The $1{,}000$-step budget at effective batch size $32$ corresponds to roughly $32{,}000$ records, which is below one full epoch over the WinDOM training partition; we deliberately do not run multiple epochs because the SFT trap (\cref{sec:finding1}) becomes more severe past one epoch.

\paragraph{GRPO hyperparameter notes.} The group size $G=8$ and the temperature $T_{\text{rollout}}=1.3$ were chosen jointly: smaller groups make $\sigma$ in \cref{eq:adv} estimate-noisy, and lower temperatures collapse the group onto a single coordinate. The completion-length cap of $64$ tokens is enough for the longest valid tool-call action our parser has accepted in any of the runs we logged. The KL coefficient $\beta$ is held at zero because, on our setup, it does not change the early-versus-saturated gap (see App.~\ref{app:negative}); the cold-start policy enters only as the initialization of $\theta$, not as a fixed reference. The reward weights ($0.1$ format $+$ peak-$1.0$ Gaussian click-in-target) are calibrated so that a parsable but spatially incorrect rollout receives reward in $[0.1, \exp(-\tfrac{1}{2})\cdot 1.0 + 0.1]\!\approx\![0.1, 0.71]$ depending on its distance from the GT center, while a parsable click at the GT center receives $1.1$ and a non-parsable rollout receives $0$. This gradient is what allows GRPO to reward proximity rather than only success.

\begin{table*}[ht]
\centering
\small
\caption{Cold-start hyperparameters (shared across Direct SFT and both SFD modes).}
\label{tab:hp-cold}
\begin{tabular}{l l}
\toprule
Hyperparameter & Value \\
\midrule
Student backbone & 2B-parameter vision-language transformer (\cref{app:setup}) \\
Optimizer & AdamW (default $\beta_1{=}0.9$, $\beta_2{=}0.999$, $\epsilon{=}10^{-8}$) \\
Learning rate & $5\times 10^{-6}$ \\
LR schedule & Cosine decay, $3\%$ linear warm-up \\
Effective batch size & $32$ (gradient accumulation as needed for GPU memory) \\
Precision & bfloat16 (mixed precision; optimizer states in fp32) \\
Total steps & $1{,}000$ (one checkpoint every $100$) \\
SFD-EMA decay $\alpha$ & $0.99$ \\
SFD distillation loss & token-level forward KL at teacher-token positions (\cref{eq:sfd}) \\
SFD KL softmax temperature & $1.0$ (no softening of teacher logits) \\
Cross-size SFD teacher & frozen $\sim$4B same-family sibling of the student \\
Hint jitter (teacher-only) & $\pm 30$ integer noise per coordinate on the $[0,1000]^2$ grid, clamped \\
Click-in-bbox filter & teacher rollout kept iff its parsed click lies inside $\mathcal{B}_{\text{GT}}$ \\
\bottomrule
\end{tabular}
\end{table*}

\begin{table*}[ht]
\centering
\small
\caption{GRPO hyperparameters (shared across all six cold-start initializations).}
\label{tab:hp-grpo}
\begin{tabular}{l l}
\toprule
Hyperparameter & Value \\
\midrule
Group size $G$ & $8$ \\
Learning rate & $2\times 10^{-6}$ \\
KL coefficient $\beta$ (toward cold-start) & $0.0$ \\
Clip range $\varepsilon$ & $0.2$ \\
Maximum completion length & $64$ tokens \\
Rollout sampling temperature $T_{\text{rollout}}$ & $1.3$ \\
Optimizer & AdamW (defaults as above) \\
Training prompts (subsample of WinDOM train) & $8{,}000$ \\
Reward & $0.1\cdot\mathbf{1}[\mathbf{o}\text{ parses}]\,+\,\exp\!\big(-\tfrac{1}{2}[(\Delta x/\sigma_x)^2{+}(\Delta y/\sigma_y)^2]\big)$ \\
$(\sigma_x,\sigma_y)$ & half the GT bbox width and height in pixels \\
Total steps & $1{,}000$ (one checkpoint every $100$) \\
Precision & bfloat16 \\
\bottomrule
\end{tabular}
\end{table*}

\section{Compute}
\label{app:compute}

All training, including the cold-start and GRPO stages, runs on a single H200 GPU. All evaluation runs on a single H100 GPU. We did not use data or pipeline parallelism at any stage. The 2B student fits comfortably in a single H100 in bfloat16 at the listed batch size; the cross-size SFD setup additionally hosts the $\sim$4B teacher in inference mode in the same GPU memory, which is the constraint that pins the choice of teacher to a sibling no larger than $\sim$4B at this student size. We do not report exact wall-clock times per run, since they vary with preemption-induced restarts; we checkpoint every $100$ steps so that a preemption costs at most one checkpoint of progress.

\section{Evaluation determinism}
\label{app:determinism}
Greedy decoding produces deterministic evaluation. Three reruns of our best configuration on OSWorld-G across different H100 GPUs returned identical accuracy ($278/510$ correct on each run). Partial-run accuracies on ScreenSpot-Pro can vary across reruns because evaluation-time sample order can differ, but the full-set accuracy is identical. We therefore attribute the variance reported in the main paper entirely to training stochasticity rather than evaluation noise.

\section{WinDOM details}
\label{app:dataset}

\paragraph{Per-scene randomization.} Layout templates are sampled uniformly from \{\texttt{single}, \texttt{corners}, \texttt{mosaic}, \texttt{cascading}, \texttt{split2}, \texttt{many\_cascading}\}. Viewports are drawn from the weighted pool $\{1920{\times}1080{:}\!\times\!6,\,2560{\times}1440{:}\!\times\!2,\,1920{\times}1200{:}\!\times\!1,\,1680{\times}1050{:}\!\times\!1\}$. Wallpapers are sampled uniformly from a curated bundle.

\paragraph{Bounding-box extraction.} The interactable ARIA roles consumed by stage (iii) of \cref{sec:dataset} are: \texttt{button}, \texttt{link}, \texttt{textbox}, \texttt{menuitem}, \texttt{checkbox}, \texttt{radio}, \texttt{tab}, \texttt{switch}, \texttt{combobox}, \texttt{option}, \texttt{treeitem}, \texttt{listitem}, and \texttt{img}.

\paragraph{Geometric cleaning.} A box is dropped if its mean pixel brightness lies outside $[10, 235]$ on a $[0, 255]$ scale (blank or saturated regions), if its area exceeds $5\%$ of the viewport (panels rather than elements), or if it would cause a trajectory to retain more than $36$ elements. Perceptual-hash deduplication suppresses near-duplicate captures across trajectories.

\paragraph{Splits.} Each trajectory identifier is bucketed by SHA-1 hash into target ratios of $85\%/5\%/10\%$ (train/val/test). Because trajectory record counts vary, the realized record-level shares are $82.9\%/6.0\%/11.2\%$ ($45{,}095/3{,}260/6{,}070$ records over $420/28/52$ trajectories). The held-out test split contains $6{,}070$ records; the in-distribution evaluation in \cref{sec:experiments} uses a fixed $2{,}000$-record subsample of that split for all comparisons.

\paragraph{Cleaned release.} A separately maintained \texttt{cleaned-v2} release applies the geometric cleaning above more aggressively and is intended for downstream users that prefer fewer, higher-precision records; the experiments in this paper do not use it.

\paragraph{Provenance flags.} Each record carries flags indicating which refinement stages it has passed (template, automatic refinement, automatic verification, vision-agent edit, human review), along with the layout template, viewport, parent scene, and developer-supplied action label. On the released $54{,}425$-record corpus the QA-pass and edit fractions are: $100\%$ (\texttt{\_qa\_full\_pass}; the QA-pass classifier is run on every kept candidate), $25.8\%$ (\texttt{\_relabel\_v2}; recovered after a Gemini-2.5-Flash relabel pass), $0.9\%$ (\texttt{\_agent\_qa\_edited}; edited by a Claude vision agent in the multi-agent QA pass), and $0.03\%$ (\texttt{\_human\_corrected}; $19$ records manually verified by a human reviewer). The keep/edit/reject ratios at the QA classifier stage are not retained per-record because rejected candidates were dropped at QA time rather than carried; we will publish the per-stage drop counts and the per-record provenance manifest with the corpus release so that downstream users can re-derive the keep/edit/reject distribution.

\section{Headline-row reproduction recipe}
\label{app:recipe}
The headline row of \cref{tab:main} (SFD-4B Early-init RL, $76.5/55.7/54.5/88.8$ on in-distribution / SS-Pro / OSG / SS-V2) is produced by the following three steps; every hyperparameter, prompt, and decoding setting referenced below is fixed in \cref{app:setup,app:hp,app:prompts}.

\begin{enumerate}[leftmargin=*,topsep=2pt,itemsep=2pt]
\item \emph{Cold-start.} Train the SFD-4B mode of \cref{sec:sfd} on the WinDOM training partition for $1{,}000$ steps under the cold-start hyperparameters of \cref{tab:hp-cold}, with the cross-size $4$B teacher receiving the $\pm 30/[0,1000]^2$ jittered ground-truth-bbox hint and the click-in-bounding-box rejection sampler of \cref{sec:sfd}. Save the student checkpoint at supervised step $100$.
\item \emph{Reinforcement learning.} Initialise GRPO from the step-$100$ checkpoint and run for $1{,}000$ steps on $8{,}000$ training prompts subsampled deterministically from the WinDOM training partition under seed $42$. Use the GRPO hyperparameters of \cref{tab:hp-grpo}, the Gaussian click-in-bounding-box reward of \cref{eq:reward}, and the group-relative advantage of \cref{eq:adv}. Save a checkpoint every $100$ GRPO steps.
\item \emph{Evaluation.} The checkpoint at GRPO step $400$ is the headline row. Decode greedily under the protocol of \cref{app:setup} and score using the click-in-target indicator of \cref{eq:cgt} on the in-distribution WinDOM split, ScreenSpot-Pro, OSWorld-G, and ScreenSpot-V2.
\end{enumerate}

The cross-seed reproducibility of this row is reported in \cref{app:variance}.

\section{System prompts and tool-call schema}
\label{app:prompts}

The student model is trained and evaluated with the same system prompt, identical to the official Qwen-VL grounding template. Every input is a (system message, user message with image + textual instruction) pair, and the model is required to emit a single \texttt{<tool\_call>}\ldots\texttt{</tool\_call>} JSON block describing a left-click. We reproduce the prompt verbatim:

\begin{quote}\small\ttfamily
You are a helpful assistant. The user will give you an instruction, and you MUST left click on the corresponding UI element via tool call. If you are not sure about where to click, guess a most likely one.\\[2pt]
\# Tools\\
You may call one or more functions \ldots resolution is \{sw\}x\{sh\}\ldots
\end{quote}

\noindent The placeholders \texttt{\{sw\}}, \texttt{\{sh\}} are filled in at runtime with the screenshot resolution. Both SFD teachers (cross-size $4$B and same-size EMA self-teacher) receive the same system message but with an additional \emph{hint} suffix appended to the user turn:

\begin{quote}\small\ttfamily
\textbackslash nHint: the target element is approximately in the region [\{hx1\}, \{hy1\}, \{hx2\}, \{hy2\}] (xmin, ymin, xmax, ymax).\textbackslash nNow answer with the tool call only:
\end{quote}

\noindent The four hint coordinates are the ground-truth bounding box on the $[0, 1000]^2$ scale, jittered by an integer in $[-30, +30]$ per coordinate and clamped to range. The student is never shown this hint.

\section{Three-seed GRPO replication}
\label{app:variance}

We launched three independent GRPO runs of our best configuration (Early-init RL on top of the SFD-4B cold-start at step 100), differing only in the seed passed to the trainer (\texttt{seed}$\in\{42, 123, 2024\}$). The seed is consumed in two places: (i) a \texttt{random.Random(seed)} that selects the $8{,}000$-prompt subsample from the WinDOM training partition, and (ii) the TRL \texttt{GRPOConfig.seed}, which Hugging Face Trainer forwards to \texttt{transformers.set\_seed} (reseeding the Python \texttt{random}, NumPy, PyTorch CPU/CUDA RNGs, and the dataloader \texttt{RandomSampler} that controls prompt-shuffle order) and to GRPO's per-step rollout sampling RNG. The cold-start checkpoint loaded as the policy initialization is byte-identical across the three runs, and greedy evaluation is deterministic, so all reported between-seed variance is contributed by what the seed actually controls inside training. \texttt{seed=42} is the configuration reported in \cref{tab:main}; all three seeds have completed evaluation across the in-distribution split, ScreenSpot-Pro, and OSWorld-G.

\textbf{Cross-seed spread on the best model.} On the best GRPO model the cross-seed accuracy range is $0.9$pp on the in-distribution split, $0.6$pp on ScreenSpot-Pro, and $0.4$pp on OSWorld-G. The cross-seed gap between SFD-4B Early-init RL and any Late-init RL or cold-start-only configuration in \cref{tab:main} therefore exceeds the seed-induced spread on every benchmark.

\section{What did not work}
\label{app:negative}
\textbf{No-cold-start GRPO at 2B.} Running GRPO directly on the base model failed: format-parse errors dominated the early reward signal and accuracy never exceeded thirty percent on the in-distribution split. At 2B scale, even a binary verifiable reward needs a warm-up. This is consistent with the larger-scale results reported in GTA1~\citep{HelloKKMe2025GTA1} and points to a scale-dependent threshold.

\textbf{Longer GRPO from saturated SFT.} Extending Late-init RL with the SFT cold-start to twice the standard length left OOD accuracy unchanged and slowly degraded in-distribution accuracy. The late-init policy does not recover with additional GRPO steps; it is locked.

\textbf{Mild KL regularization.} Adding a small KL term toward the cold-start policy ($\beta=0.01$) changed the early-versus-saturated gap by less than one point on ScreenSpot-Pro. The transfer gap originates in the cold-start, not in KL drift during reinforcement learning.

\end{document}